\documentclass[journal]{IEEEtran}
\usepackage{cite}
\usepackage{amsmath,amssymb,amsfonts}
\usepackage{algorithmic}
\usepackage{graphicx}
\usepackage{textcomp}
\usepackage{siunitx}
\usepackage{enumerate}
\usepackage{amsmath}
\usepackage{amssymb}
\usepackage{algorithm,algorithmic}
\usepackage{graphicx}
\usepackage{upgreek}
\usepackage{bm}
\usepackage{blindtext}
\usepackage{tikz}
\usepackage{subfig}
\usepackage[export]{adjustbox}
\usepackage{dblfloatfix} 
\usepackage{comment}
\usepackage{nicefrac}
\usepackage{accents}
\usetikzlibrary{shapes,arrows,positioning,calc}

\newcommand{\vect}[1]{\boldsymbol{\mathbf{#1}}}
\newcommand{\diff}{d}

\DeclareMathAlphabet{\mathpzc}{OT1}{pzc}{m}{it}
\def\realR{\mathbb{R}}
\newcommand{\ie}{{\em i.e.}}
\newcommand{\eg}{{\em e.g.}}
\usepackage{flushend}
\usepackage{cite}

\makeatletter
\let\NAT@parse\undefined
\makeatother
\usepackage{hyperref}  %hyperref still needs to be put at the end!

\tikzset{
	block/.style = {draw, fill=white, rectangle, minimum height=3em, minimum width=3em},
	lpfblock/.style = {draw, fill=white, rectangle, minimum height=2em, minimum width=3em},
	smallblock/.style = {draw, fill=white, rectangle, minimum height=2em, minimum width=2em},
	lpfblocktall/.style = {draw, fill=white, rectangle, minimum height=3em, minimum width=3em},
	bigblock/.style = {draw, fill=white, rectangle, minimum height=11em, minimum width=3em},
	uavbigblock/.style = {draw, fill=white, rectangle, minimum height=13em, minimum width=3em},
	bigblock2/.style = {draw, fill=white, rectangle, minimum height=12em, minimum width=6em},
	bigblock3/.style = {draw, fill=white, rectangle, minimum height=14em, minimum width=6em},
	concat/.style =  {draw, fill=black, rectangle, minimum height=4em, minimum width=0.25em,inner sep = 0em},
	tmp/.style  = {coordinate}, 
	sum/.style= {draw, fill=white, circle, node distance=3em},
	gain/.style = {regular polygon, regular polygon sides=3,
		draw, fill=white, text width=1em,
		inner sep=0.15em, outer sep=0em,
		shape border rotate=-90,minimum height=3em, minimum width=3em},
	smallgain/.style = {regular polygon, regular polygon sides=3,
		draw, fill=white, text width=1em,
		inner sep=0.em, outer sep=0mm,
		shape border rotate=-90,minimum height=1.5em, minimum width=1.5em},
	smallishgain/.style = {regular polygon, regular polygon sides=3,
		draw, fill=white, text width=1em,
		inner sep=0.07em, outer sep=0mm,
		shape border rotate=-90,minimum height=2em, minimum width=2em},
	midgain/.style = {regular polygon, regular polygon sides=3,
		draw, fill=white, text width=1em,
		inner sep=0.em, outer sep=0mm,
		shape border rotate=-90,minimum height=3em, minimum width=3em},
	midpgain/.style = {regular polygon, regular polygon sides=3,
		draw, fill=white, text width=1em,
		inner sep=0.em, outer sep=0mm,
		shape border rotate=-90,minimum height=3.85em, minimum width=3.85em},
	biggain/.style = {regular polygon, regular polygon sides=3,
		draw, fill=white, text width=1em,
		inner sep=0.em, outer sep=0mm,
		shape border rotate=-90,minimum height=4.5em, minimum width=4.5em},
	gainleft/.style = {regular polygon, regular polygon sides=3,
		draw, fill=white, text width=1em,
		inner sep=0.2em, outer sep=0mm,
		shape border rotate=90,minimum height=3em, minimum width=3em},
	smallconcat/.style =  {draw, fill=black, rectangle, minimum height=3em, minimum width=0.25em,inner sep = 0em},
	input/.style = {coordinate},
	output/.style= {coordinate},
	pinstyle/.style = {pin edge={to-,thin,black}
	},
	crossline/.style={preaction={draw=white, -,line width=2pt}}
}

\definecolor{myblue}{rgb}{0.61, 0.87, 1.0}
\definecolor{mygreen}{rgb}{0.7, 0.93, 0.36}
\definecolor{myred}{rgb}{1.0, 0.71, 0.76}
\definecolor{myyellow}{rgb}{0.98, 0.93, 0.37}
\definecolor{myorange}{rgb}{1.0, 0.74, 0.53}

\begin{document}
\title{Accurate Tracking of Aggressive Quadrotor Trajectories using Incremental Nonlinear Dynamic Inversion and Differential Flatness}
\author{Ezra Tal, \IEEEmembership{Student Member, IEEE}, Sertac Karaman, \IEEEmembership{Member, IEEE}
\thanks{This work was partly supported by Office of Naval Research (ONR) grant N00014-17-1-2670. A preliminary version of this article was presented at 57th IEEE Conference on Decision and Control (CDC 2018) \cite{TalKaraman-CDC2018}.}
\thanks{E. Tal and S. Karaman are with the Department of Aeronautics and Astronautics and the Laboratory for Information and Decision Systems, Massachusetts Institute of Technology (MIT), Cambridge, MA 02139, USA.
	(e-mail: $\{$eatal,sertac$\}$@mit.edu)}}
\maketitle

\begin{abstract}
Autonomous unmanned aerial vehicles (UAVs) that can execute aggressive (\ie, high-speed and high-acceleration) maneuvers have attracted significant attention in the past few years. 
This paper focuses on accurate tracking of aggressive quadcopter trajectories. We propose a novel control law for tracking of position and yaw angle and their derivatives of up to fourth order, specifically, velocity, acceleration, jerk, and snap along with yaw rate and yaw acceleration. 
Jerk and snap are tracked using feedforward inputs for angular rate and angular acceleration based on the {\em differential flatness} of the quadcopter dynamics.
Snap tracking requires direct control of body torque, which we achieve using closed-loop motor speed control based on measurements from optical encoders attached to the motors. 
The controller utilizes {\em incremental nonlinear dynamic inversion (INDI)} for robust tracking of linear and angular accelerations despite external disturbances, such as aerodynamic drag forces. Hence, prior modeling of aerodynamic effects is not required.
We rigorously analyze the proposed control law through response analysis, and we demonstrate it in experiments. 
The controller enables a quadcopter UAV to track complex 3D trajectories, reaching speeds up to 12.9 m/s and accelerations up to 2.1g, while keeping the root-mean-square tracking error down to 6.6 cm, in a flight volume that is roughly 18 m by 7 m and 3 m tall.
We also demonstrate the robustness of the controller by attaching a drag plate to the UAV in flight tests and by pulling on the UAV with a rope during hover.
\end{abstract}


	\section*{Supplemental Material} 
%	\vspace{-0.05in}

A video of the experiments can be found at \url{https://youtu.be/K15lNBAKDCs}.

\section{Introduction}\label{sec:intro}
\IEEEPARstart{H}{igh-speed} aerial navigation through complex environments has been a focus of control theory and robotics research for decades. 
More recently, {\em drone racing}, in which remotely-operated rotary-wing aircraft are piloted through challenging, obstacle-rich courses at very high speeds, has further inspired and popularized this research direction.
Development of fully-autonomous drone racers requires accurate control of aircraft during aggressive, \ie, high-speed and agile, maneuvers. At high speeds, aerodynamic drag, which is hard to model, becomes a dominant factor. This poses an important challenge in control design. Additionally, accurate tracking of a reference trajectory with fast-changing acceleration requires considering its higher-order time derivatives, \ie, jerk and snap. In contrast, control design for rotary-wing, vertical take-off and landing (VTOL) aircraft at low speeds typically neglects both aerodynamics and higher-order derivatives.

\begin{figure}
	\centering
	\includegraphics[width=\linewidth]{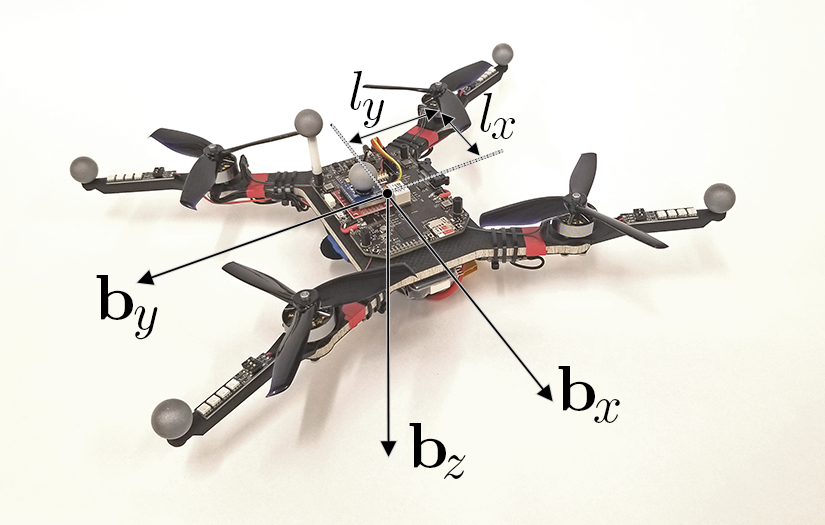}
	\caption{Quadrotor with body-fixed reference system and moment arm definitions.}\label{fig:refsystems}
	\vspace{-0.3in}
	
\end{figure}

In this paper, we propose a novel control design for accurate tracking of aggressive trajectories using a quadcopter aircraft, such as the one shown in Fig.~\ref{fig:refsystems}.
The proposed controller generates feedforward control inputs based on differential flatness of the quadcopter dynamics, and uses incremental nonlinear dynamic inversion (INDI) to handle external disturbances, such as aerodynamic drag.

Nonlinear dynamic inversion (NDI), also called feedback linearization, enables the use of a linear control law by transforming the nonlinear dynamics into a linear input-output map~\cite{jjslotine,isidori,sastry2013nonlinear}.
Although variants of NDI were quickly developed for flight control \cite{snell1992nonlinear,bugajski1992nonlinear,hauser1992nonlinear,enns1994dynamic,koo1998output}, it is well known that exact dynamic inversion inherently suffers from lack of robustness \cite{lee2009feedback}.
As a result, other nonlinear control methods, \eg, adaptive sliding mode \cite{xu2006sliding,madani2006backstepping,lee2009feedback} and backstepping designs \cite{frazzoli2000trajectory}, have been considered in order to achieve robustness in flight control.
More recently, an incremental version of nonlinear dynamic inversion has been developed \cite{indi,simplicio2013acceleration}, based on earlier derivations \cite{smith1998simplified,bacon2000reconfigurable}, which provide robustness by incrementally applying control inputs based on inertial measurements.
In existing literature, the INDI technique has been applied to quadcopters for stabilization, \eg, for robust hovering \cite{smeur2015adaptive,smeur2017cascaded}, but not for trajectory tracking.

Differential flatness, or feedback linearizability, of a dynamics system allows expressing all state and input variables in terms of a set of flat outputs and its derivatives~\cite{fliess1992lessystemesnon,martin1992contribution,van1998differential,martin2003flat,fliessintro}.
In the context of flight control, this property enables reformulation of the trajectory tracking problem as a state tracking problem~\cite{koo1998output,martin1996aircraft}.
Specifically, it enables consideration of higher-order derivatives of the reference trajectory through feedforward state and input references, which has also been applied to control~\cite{rivera2010flatness,ferrin2011differential,engelhardt2016flatness,faessler2018differential}.

Quadcopter aircraft are relatively easy to maneuver and experiment with. Arguably, these qualities make them ideal for drone racing events. For the same reasons, they have been heavily used as experimental platforms in robotics and control theory research since the start of this century \cite{starmac,kroo2000mesicopter,pounds2002design,hamel2002dynamic}.
Complex trajectory tracking control systems have been designed and demonstrated for aircraft in motion capture rooms, where the position and the orientation of the aircraft can be obtained with high accuracy~ \cite{valenti2006indoor,how2008real,zarovy2010experimental,ducard2009autonomous,michael2010grasp,ol2008flight,bieniawski2008micro}. Agile maneuvers for quadcopter aircraft have also been demonstrated~\cite{mellinger2012trajectory,muller2011quadrocopter}. Despite being impressive, these demonstrations have showcased complex trajectories only at relatively slow speeds, \eg, less than 2 \si{m/s}, so that aerodynamic forces and moments may be neglected.
At higher speeds, aerodynamic effects heavily influence the vehicle dynamics.
This has been addressed in recent research through modeling \cite{faessler2018differential,svacha2017improving,kai2017nonlinear}, estimation \cite{liu2016robust,huang2009aerodynamics}, and learning \cite{schoellig2012optimization} of aerodynamic drag effects towards tracking control in high-speed flight.

The main contribution of this paper is a trajectory tracking control design that achieves accurate tracking during high-speed and high-acceleration maneuvers without depending on modeling or estimation of aerodynamic drag parameters.
The design exploits differential flatness of the quadcopter dynamics to generate feedforward control terms based on the reference trajectory and its derivatives up to fourth order, i.e., velocity, acceleration, jerk, and snap. Modeling inaccuracies and disturbances due to aerodynamic drag are compensated for using incremental control based on the INDI technique.
This control design is novel in the following ways.
Firstly, the design incorporates direct tracking of reference snap through accurate control of the motor speeds using optical encoders attached to each motor.
We recognize that snap is directly related to vehicle angular acceleration, and thus to the control torque acting on the quadcopter.
Accurate application of torque commands is achieved by precise closed-loop control of the motor speeds using measurements from the optical encoders.
To the best of our knowledge, the direct control over snap using motor speed measurements is novel.
In contrast, trajectory tracking control based on body rate inputs, \eg, using a typical inner-loop flight controller, is incapable of truly considering reference snap.
Secondly, we develop a novel INDI control design for quadcopter trajectory tracking.
Thrust and torque commands are applied incrementally for robustness against significant external disturbances, such as aerodynamic drag, without the need to model or estimate said disturbances.
As far as we are aware, the proposed controller is the first design that is tailored for trajectory tracking, as existing INDI flight control designs focus on state regulation, \eg, for maintaining hover under external disturbances.
Thirdly, we provide and evaluate a novel implementation of INDI angular acceleration control that includes nonlinear computation of the control increments, as opposed to the existing implementations that use inversion of linearized control effectiveness equations.
Finally, we demonstrate the proposed controller in experiments, and we rigorously analyze the benefits of the key aspects of our control design through response analysis. 
In our experiments, the proposed control law enables a unmanned aerial vehicle (UAV) to track complex 3D trajectories, reaching speeds up to 12.9 \si{m/s} and accelerations up to 2.1\si{g}, while keeping the root-mean-square (RMS) tracking error down to 6.6 \si{cm}, in a flight volume that is roughly 18 \si{m} long, 7 \si{m} wide, and \si{3} m tall. 
We also demonstrate the robustness of the controller by attaching a drag plate to the UAV in flight tests and by pulling on the UAV using a tensioned wire during hover. The improved performance due to the tracking of reference jerk and snap through feedforward angular velocity and angular acceleration inputs is also demonstrated both in theoretical analysis and in experiments.

A preliminary version of this paper was previously presented at 57th IEEE Conference on Decision and Control (CDC 2018) \cite{TalKaraman-CDC2018}.
The significant extensions introduced in the current work include a reformulation of the controller using quaternion attitude representation, a more elaborate description of its architecture, the response analysis section in its entirety, and new experimental results at increased speeds.
The application of the singularity-free quaternion representation enables tracking of very aggressive trajectories that would incur singular states if an Euler angle representation were used.

The paper is structured as follows: Nomenclature is presented in Table \ref{tab:nomenclature}. In Section \ref{sec:prelims}, the quadrotor model is specified, and we show how differential flatness is used to formulate feedforward control inputs in terms of the reference trajectory. In Section \ref{sec:trajtrack}, we describe the architecture of the trajectory tracking controller, and its individual components. Analysis in Section \ref{sec:analysis} illustrates the robustness of INDI and the effect of the feedforward control inputs through response analysis. Finally, we give experimental results from real-life flights in Section \ref{sec:experiments}.

\begin{table*}
	\centering
	\caption{%
		Nomenclature. The subscript $ref$ is used to indicate elements of the reference trajectory function and its time derivatives, as well as feedforward variables directly obtained from the reference trajectory function.
		The subscript $c$ is used for commanded values that are obtained from a feedback control loop.
		Low-pass filtered measurements and signals obtained from such measurements are indicated by the subscript $f$.}
	\label{tab:nomenclature}
	\hrulefill
	
	\vspace{0.15em}
	\begin{minipage}[b]{0.49\textwidth}
		\begin{tabular}{ll}
			$\circ$ & Hamilton quaternion product\\
			$\bullet^{\circ n}$&$n$-th Hadamard (element-wise) power\\
			$\left[\bullet\right]_\times$ & cross-product matrix\\
			$\vect{a}$& linear acceleration in inertial frame, \si{m/s\textsuperscript{2}}\\
			$\vect{a}^b$& linear acceleration including gravitational acceleration\\
			& in body-fixed frame, \ie, as measured by IMU, \si{m/s\textsuperscript{2}}\\
			$\vect{b}_x$, $\vect{b}_y$, $\vect{b}_z$& basis vectors of body-fixed frame\\
			$C^n$ & $n$-th order differentiability class\\
			$\vect{f}_{ext}$& external disturbance force vector in inertial frame, \si{N}\\
			$g$& gravitational acceleration, \si{m/s\textsuperscript{2}}\\
			$\vect{G}_1$& propeller speed control effectiveness matrix\\
			$\vect{G}_2$& propeller acceleration control effectiveness matrix\\
			$H(s)$ & low-pass filter transfer function\\
			$\vect{i}_x$, $\vect{i}_y$, $\vect{i}_z$& standard basis vectors\\
			$\vect{j}$& jerk in inertial frame, \si{m/s\textsuperscript{3}}\\
			$\vect{J}$& vehicle moment of inertia matrix, \si{kg}$\cdot$\si{m\textsuperscript{2}}\\
			$J_{yy}$& vehicle moment of inertia around $\vect{b}_y$-axis, \si{kg}$\cdot$\si{m\textsuperscript{2}}\\
			$J_{r_z}$& motor rotor and propeller moment of inertia, \si{kg}$\cdot$\si{m\textsuperscript{2}} \\
			$k_\theta$, $k_{q}$& scalar control gains\\
			$k_G$& linearized pitch control effectiveness, \si{kg}$\cdot$\si{m\textsuperscript{2}/(rad}$\cdot$\si{s)}\\
			$k_{\mu_z}$& propeller torque coefficient, \si{kg}$\cdot$\si{m\textsuperscript{2}/rad\textsuperscript{2}}\\
			$k_\tau$& propeller thrust coefficient, \si{kg}$\cdot$\si{m/rad\textsuperscript{2}}\\
			$\vect{K_x}$, $\vect{K_v}$,&diagonal control gain matrices\\
			$\vect{K_a}$, $\vect{K_\xi}$&\\
			$\vect{K_{\Omega}}$, $\vect{K_{I_\omega}}$&\\
			$l_x$& moment arm component parallel to $\vect{b}_x$-axis, \si{m}\\
			$l_y$& moment arm component parallel to $\vect{b}_y$-axis, \si{m}\\
			$m$& vehicle mass, \si{kg}\\
			$M(s)$& motor (control) dynamics transfer function\\
			$NI$ & transfer function corresponding to non-incremental controller
		\end{tabular}
	\end{minipage} \hfill
	\begin{minipage}[b]{0.49\textwidth}
		\begin{tabular}{ll}
			$\mathpzc{p}$& polynomial relating motor speeds to throttle inputs\\
			$q$& vehicle pitch rate around $\vect{b}_y$-axis, \si{rad/s}\\
			$\vect{r}_\psi$& yaw direction vector in inertial frame\\
			$\vect{R}$& body-fixed to inertial frame rotation matrix\\
			${s}$& Laplace variable\\
			$\vect{s}$& snap in inertial frame, \si{m/s\textsuperscript{4}}\\
			$\vect{S}$&angular rate to yaw rate transformation\\
			$SO(3)$& three-dimensional special orthogonal group\\
			$t$ & time, \si{s}\\
			$T$& thrust, \si{N}\\
			$\mathbb{T}$ & circle group\\
			$\vect{v}$& velocity in inertial frame, \si{m/s}\\
			$\vect{x}$& position in inertial frame, \si{m}\\
			$\alpha$& vehicle pitch acceleration around $\vect{b}_y$-axis, \si{rad/s\textsuperscript{2}}\\
			$\Delta$& modeling error parameter\\
			$\vect{\zeta}$ & throttle command vector\\
			$\theta$& vehicle pitch angle, \si{rad} \\
			$\vect{\mu}$& control moment vector, \si{N}$\cdot$\si{m}\\
			$\vect{\mu}_{ext}$& external disturbance moment vector, \si{N}$\cdot$\si{m}\\
			$\vect{\xi}$&normed quaternion attitude vector\\
			$\xi^w$, $\xi^x$, $\xi^y$, $\xi^z$&elements of $\vect{\xi}$\\
			$\vect{\xi}_c$&incremental command relative to current attitude\\
			$\vect{\xi}_e$&vector of error angles in body-fixed frame\\
			$\vect{\sigma}_{ref}(t)$& reference trajectory function, \si{m}, \si{rad}\\
			$\tau$& specific thrust, \si{m/s\textsuperscript{2}}\\
			$\uptau_m$& motor dynamics time constant, \si{s}\\
			$\psi$& vehicle yaw angle, \si{rad}\\
			$\omega$&deviation from hover state motor rotation speed, \si{rad/s}\\
			$\omega_0$&hover state motor rotation speed, \si{rad/s}\\
			$\vect{\omega}$& vector of four motor rotation speeds, \si{rad/s}\\
			$\vect{\Omega}$& vehicle angular velocity in body-fixed frame, \si{rad/s}
		\end{tabular}
		\vspace{-17.4em}
	\end{minipage}
	\vspace{0.3em}
	
	\hrulefill
\end{table*}

\section{Preliminaries}\label{sec:prelims}
In this section, we describe the quadrotor dynamics model, and its differential flatness property.
Specifically, we show how the control system utilizes this property to track the reference trajectory jerk and snap through feedforward angular rate and angular acceleration inputs.

\subsection{Quadrotor Model}
We consider a 6 degree-of-freedom (DOF) quadrotor, as shown in Fig. \ref{fig:refsystems}. The unit vectors depicted in the figure are the basis of the body-fixed reference frame and form the rotation matrix $\vect{R}=[
\vect{b}_x\;\vect{b}_y\;\vect{b}_z
] \in SO(3)$, which gives the transformation from the body-fixed reference frame to the inertial reference frame.
The basis of the north-east-down (NED) inertial reference frame consists of the columns of the identity matrix $[
\vect{i}_x\;\vect{i}_y\;\vect{i}_z]$.

The vehicle translational dynamics are given by
\begin{align}
\vect{\dot x} &= \vect{v},\label{eq:xdot}\\
\vect{\dot v} &= g\vect{i}_z + \tau \vect{b}_z + m^{-1}\vect{f}_{ext},\label{eq:vdot}
\end{align}
where $\vect{x}$ and $\vect{v}$ are the position and velocity in the inertial reference frame, respectively. Equation \eqref{eq:vdot} includes three contributions to the linear acceleration. Firstly, the gravitational acceleration $g$ in downward direction. Secondly, the specific thrust ${\tau}$, which is the ratio of the total thrust $T$ and the vehicle mass $m$. Note that the thrust vector is always aligned with the $\vect{b}_z$-axis, so that the quadrotor must pitch or roll to accelerate forward, backward or sideways. Finally, the external disturbance force vector $\vect{f}_{ext}$ accounts for all other forces acting on the vehicle, such as aerodynamic drag.

The rotational dynamics are given by
\begin{align}
\vect{\dot \xi} &= \frac{1}{2}\vect{\xi}\circ\vect{\Omega},\label{eq:xidot}\\
\vect{\dot \Omega} &= \vect{J}^{-1}(\vect{\mu} + \vect{\mu}_{ext}-\vect{\Omega} \times \vect{J}\vect{\Omega}), \label{eq:Omegadot}
\end{align}
where
$\vect{\Omega}$ is the angular velocity in the body-fixed reference frame, and
$\vect{\xi}= [\xi^w \; \xi^x \; \xi^y \; \xi^z]^T$ is the normed quaternion attitude vector, so that $\vect{R}\vect{x} = \vect{\xi}\circ\vect{x}\circ\vect{\xi}^{-1}$ with $\circ$ the Hamilton product.
Note that a zero magnitude element is implied when multiplying three-element vectors with quaternions.
The matrix $\vect{J}$ is the vehicle moment of inertia tensor.
The control moment vector is indicated by $\vect{\mu}$, and the external disturbance moment vector by $\vect{\mu}_{ext}$. The third term of \eqref{eq:Omegadot} accounts for the conservation of angular momentum.

Each propeller axis is assumed to be aligned perfectly with the $\vect{b}_z$-axis, so that all motor speeds are described by the four-element vector $\vect{\omega} > 0$.
The total thrust $T$ and control moment vector in body-reference frame $\vect{\mu}$ are given by
\begin{equation}\label{eq:mutau}
\left[\begin{array}{c}
\vect{\mu}\\
T
\end{array}\right]=\vect{G}_1 \vect{\omega}^{\circ 2} + \vect{G}_2 {\vect{\dot\omega}},
\end{equation}
where $^\circ$ indicates the Hadamard power;
\begin{equation}\label{eq:G1}
\vect{G}_1=\left[\begin{array}{cccc}
l_yk_\tau&-l_yk_\tau&-l_yk_\tau&l_yk_\tau\\
l_xk_\tau&l_xk_\tau&-l_xk_\tau&-l_xk_\tau\\
-k_{\mu_z}&k_{\mu_z}&-k_{\mu_z}&k_{\mu_z}\\
-k_\tau&-k_\tau&-k_\tau&-k_\tau
\end{array}\right],
\end{equation}
with $l_x$ and $l_y$ the moment arms indicated in Fig. \ref{fig:refsystems}, $k_\tau$ the propeller thrust coefficient, and $k_{\mu_z}$ the propeller torque coefficient; and
\begin{equation}\label{eq:G2}
\vect{G}_2 = \left[\begin{array}{cccc}
0&0&0&0\\
0&0&0&0\\
-J_{r_z}&J_{r_z}&-J_{r_z}&J_{r_z}\\
0&0&0&0\\
\end{array}\right]
\end{equation}
with $J_{r_z}$ the rotor and propeller moment of inertia.
The second term in \eqref{eq:mutau} represents the control torque directly due to motor torques.
Due to their relatively small moment of inertia, the contribution of the motors to the total vehicle angular momentum may be neglected.

\subsection{Differential Flatness}\label{sec:diffflatness}
The controller aims to accurately track the reference trajectory defined by the following function:
\begin{equation}\label{eq:traj}
\vect{\sigma}_{ref}(t)=[\vect{x}_{ref}(t)^T\;\psi_{ref}(t)]^T,
\end{equation}
which consists of four differentially flat outputs, \ie, the quadrotor position in the inertial reference frame $\vect{x}_{ref}(t)\in \realR^3$, and the vehicle yaw angle $\psi_{ref}(t)\in \mathbb{T}$, where $\mathbb{T}$ denotes the circle group. Henceforward, we do not explicitly write the time argument $t$ everywhere.

For \eqref{eq:traj} to be dynamically feasible, it is required that $\vect{x}_{ref}$ is of differentiability class $C^4$, \ie, its first four derivatives exist and are continuous, and that $\psi_{ref}$ is of class $C^2$. The temporal derivatives of $\vect{x}_{ref}$ are successively the reference velocity $\vect{v}_{ref}$, the reference acceleration $\vect{a}_{ref}$, the reference jerk $\vect{j}_{ref}$, and the reference snap $\vect{s}_{ref}$, all in the inertial reference frame. Similarly, temporal differentiation of $\psi_{ref}$ gives the yaw rate $\dot \psi_{ref}$, and the yaw acceleration $\ddot \psi_{ref}$.

The quadcopter dynamics are differentially flat, so that we can express its states and inputs as a function of $\vect{\sigma}_{ref}(t)$ and its derivatives.
This enables reformulation of the trajectory tracking problem as a state tracking problem.
In this section, we derive expressions for the angular rate reference $\vect{ \Omega}_{ref}$, and the angular acceleration reference $\vect{\dot \Omega}_{ref}$ in terms of trajectory jerk, snap, yaw rate, and yaw acceleration.
These reference states will be applied as feedforward inputs in the trajectory tracking control design.

Taking the derivative of \eqref{eq:vdot} yields the following expression for jerk:
\begin{equation}
\vect{j} = \tau \vect{R}\left[\vect{i}_z\right]^T_\times \vect{\Omega}+\dot \tau \vect{b}_z\label{eq:phithetadot},
\end{equation}
where $\left[\bullet\right]_\times$ indicates the cross-product matrix, and variations in the unmodeled external force $\vect{f}_{ext}$ are neglected.
This external force consists chiefly of body drag and rotor drag~\cite{hoffmann2007quadrotor,martin2010true}.
Both contributions can be included in the differential flatness transform~\cite{faessler2018differential}, but the resulting controller will depend on a vehicle-specific aerodynamics model.
Instead, we forgo modeling of the external force and instead use sensor-based control to directly compensate for it.
Therefore, our controller is able to handle external disturbances without depending on a vehicle-specific model, as described in the next section.

By taking the derivative once more, the following expression for snap is found:
\begin{equation}\label{eq:snap}
\vect{s} = \vect{R}\left(\ddot \tau \vect{i}_z +  (2\dot \tau+ \tau \left[\vect{\Omega}\right]_\times) \left[\vect{i}_z\right]^T_\times \vect{ \Omega} + \tau\left[\vect{i}_z\right]^T_\times  \vect{\dot\Omega}\right).
\end{equation}
According to typical aerospace convention, we define yaw as the angle between $\vect{i}_x$ and the vector
\begin{equation}\label{eq:rpsi1}
\vect{r}_\psi = \left[\begin{array}{ccc}
b_x^1 & b_x^2 & 0
\end{array}\right]^T
\end{equation}
with superscripts indicating individual elements of $\vect{b}_x$.
Taking the derivative of \eqref{eq:rpsi1} using $\vect{\dot R} = \vect{R}[\vect{\Omega}]_\times$, we obtain the following expression for the yaw rate:
\begin{equation}\label{eq:df_psidot}
\dot \psi = \frac{\vect{r}_\psi \times \vect{\dot r}_\psi}{\vect{r}_\psi^T \vect{r}_\psi} = \underbrace{\frac{\left[\begin{array}{cc}
	-b_x^2 & b_x^1
	\end{array}\right]}{\vect{r}_\psi^T \vect{r}_\psi}\left[\begin{array}{ccc}
0 &-b_z^1 & b_y^1\\
0 &-b_z^2 & b_y^2
\end{array}\right]}_{\vect{S}}\vect{\Omega} = \vect{S}\vect{\Omega},
\end{equation}
and, by the product rule,  the following expression for the yaw acceleration:
\begin{equation}
	\ddot \psi = \vect{S}\vect{\dot \Omega} +  \vect{\dot S}\vect{\Omega}.\label{eq:df_psiddot}
\end{equation}
An expression for the derivative $\vect{\dot S}$ is omitted here for brevity, but can be obtained by applying the product rule to the expression for $\vect{S}$ given in \eqref{eq:df_psidot}.
From \eqref{eq:phithetadot} and \eqref{eq:df_psidot}, we obtain the angular rate reference
\begin{equation}\label{eq:angrateref}
\left[\begin{array}{c}
\vect{\Omega}_{ref}\\
\dot \tau_{ref}
\end{array}\right] = \left[\begin{array}{cc}
\tau \vect{R}[\vect{i}_z]_\times^T & \vect{b}_z\\
\vect{S} & 0
\end{array}\right]^{-1}\left[\begin{array}{c}
\vect{j}_{ref}\\
\dot \psi_{ref}
\end{array}\right],
\end{equation}
and from \eqref{eq:snap} and \eqref{eq:df_psiddot} the angular acceleration reference
\begin{multline}\label{eq:angaccref}
\left[\begin{array}{c}
\vect{\dot\Omega}_{ref}\\
\ddot \tau_{ref}
\end{array}\right] = \left[\begin{array}{cc}
\tau \vect{R}[\vect{i}_z]_\times^T & \vect{b}_z\\
\vect{S} & 0
\end{array}\right]^{-1}\\\left(\left[\begin{array}{c}
\vect{s}_{ref}\\
\ddot \psi_{ref}
\end{array}\right] - \left[\begin{array}{c}
\vect{R}(2\dot \tau + \tau [\vect{\Omega}]_\times)[\vect{i}_z]_\times^T\vect{\Omega}\\
\vect{\dot S}\vect{\Omega}
\end{array}\right]\right).
\end{multline}
Note that these expressions also contain reference signals for the first and second derivatives of specific thrust.
However, as we are unable to command the corresponding first and second derivatives of the motor speed, these references remain unused by the controller.

\begin{table*}[!ht]
	\centering
	\caption{Overview of trajectory tracking controller components.}
	\label{tab:control_methods}
	\begin{tabular}{llllll}
		\hline
		Component & Methodology & Reference & Control Output & Description\\
		\hline
		Position and Velocity Control & PD & $\vect{x}_{ref}$, $\vect{v}_{ref}$, $\vect{a}_{ref}$ & $\vect{a}_c$ & Section \ref{sec:trajtrackacc}\\
		Linear Acceleration and Yaw Control & INDI & $\vect{a}_{c}$, $\psi_{ref}$ & $\vect{\xi}_{c}$, $T_c$ & Section \ref{sec:indilinacc}\\
		Jerk and Snap Tracking & Differential Flatness & $\vect{j}_{ref}$, $\vect{s}_{ref}$, $\dot \psi_{ref}$, $\ddot \psi_{ref}$  & $\vect{ \Omega}_{ref}$, $\vect{\dot \Omega}_{ref}$& Section \ref{sec:diffflatness}\\
		Attitude and Angular Rate Control & PD & $\vect{\xi}_{c}$, $\vect{ \Omega}_{ref}$, $\vect{\dot \Omega}_{ref}$ & $\vect{\dot \Omega}_c$ & Section \ref{sec:trajtrack2}\\
		Angular Acceleration Control & INDI & $\vect{\dot \Omega}_c$ & $\vect{\mu}_c$ & Section \ref{sec:attacccontrl}\\
		Moment and Thrust Control & Inversion & $\vect{\mu}_{c}$, $T_{c}$ & $\vect{\omega}_c$ & Section \ref{sec:motorspeedctrl}\\
		Motor Speed Control & Integrative & $\vect{\omega}_c$ & $\vect{\zeta}$ & Section \ref{sec:motorspeedctrl}\\
		\hline
	\end{tabular}
\end{table*}

\section{Trajectory Tracking Control}\label{sec:trajtrack}
The control design consists of several components based on various control methods. Table \ref{tab:control_methods} gives an overview of the components with their respective methodology, references, and control outputs. The control architecture is visualized in three block diagrams.
Figure \ref{fig:diagram2} shows the outer-loop position and velocity controller as described in Section \ref{sec:trajtrackacc}.
The intermediate control loop shown in Fig. \ref{fig:diagram1} controls linear acceleration, attitude and angular rate, and angular acceleration as described in Section \ref{sec:indilinacc}, \ref{sec:trajtrack2}, and \ref{sec:attacccontrl}, respectively. Finally, vehicle moment and thrust are directly controlled through closed-loop motor speed control in the inner loop, shown in Fig. \ref{fig:diagram3} and described in Section \ref{sec:motorspeedctrl}.

\begin{figure*}
\centering
\scriptsize
{\begin{tikzpicture}[auto, node distance=6em,>=latex']
\fill [myblue,opacity=1.0] (2.3em,-13.5em) rectangle ++(19.5em,12.5em);
\node [tmp](refin){};
\node [tmp,below of=refin,node distance=3em](xrefin){};
\node [sum,right of =xrefin,node distance = 4 em](sumx){};
\node [tmp,right of=sumx, node distance = 4em](oldspotKx){};
\node [gain,right of=sumx, node distance = 5em](Kx){$\vect{K_x}$};
\node [sum,below of = oldspotKx, node distance = 3em](sumv){};
\node [tmp,below of = xrefin, node distance = 3em](vin){};
\node [tmp,right of=sumv, node distance = 4em](oldspotKv){};
\node [gain,right of=sumv, node distance = 5em](Kv){$\vect{K_v}$};
\node [tmp,below of = vin, node distance = 3em](ain){};
\node [sum,below of = oldspotKv, node distance = 5.5em](suma1){};
\node [gain,right of=suma1, node distance = 4em](Ka){$\vect{K_a}$};
\node [sum,right of=oldspotKv, node distance = 8em](sumv2){};
\node [sum](suma2)at(ain-|sumv2){};
\node [sum,right of=oldspotKx, node distance = 12em](sumx2){};
\node [bigblock2,above right = -6 em and 12 em of oldspotKv,text width = 8 em,align=center,minimum height = 14 em](inner){\centering Acceleration and Attitude Control};

\node[tmp,below right = -7em and 0em of inner](midout){};
\node[tmp,below of=midout,node distance = 4 em](lowout){};
\node[tmp,above of = midout,node distance = 4 em](hiout){};

\node[tmp,below right = 2em and 2 em of inner](acorner){};
\node[tmp,below right = 1.5em and 2 em of acorner](vcorner){};
\node[tmp,below right = 1.5em and 2 em of vcorner](xcorner){};

\node [tmp,below left = 0em and 0em of inner](leftinner){};
\node [tmp] (in1) at (sumx2 -| leftinner){};
\node [tmp,below of = oldspotKv, node distance = 3em](midwayain){};

\draw[-] (hiout) -| (xcorner)
node [above,pos=0.1]{$\vect{x}$};
\draw[-] (midout) -| (vcorner)
node [above,pos=0.14]{$\vect{v}$};
\draw[-] (lowout) -| (acorner)
node [above,pos=0.3]{$\vect{a}_f$};

\draw[->] (acorner) -| (suma1)
node [above,pos=0.2]{$\vect{a}_f$}
node [left,pos = 0.92]{$-$};

\draw[->] (vcorner) -| (sumv)
node [above,pos=0.25]{$\vect{v}$}
node [right,pos = 0.97]{$-$};
\draw[->] (xcorner) -| (sumx)
node [above,pos=0.28]{$\vect{x}$}
node [right,pos = 0.983]{$-$};

\draw[->] (sumx)--(Kx);
\draw[->] (sumv)--(Kv);
\draw[->] (suma1)--(Ka);

\draw[->] (Kx)--(sumx2)
node [above,pos=0.94]{$+$};
\draw[->] (Kv)--(sumv2)
node [above,pos=0.9]{$+$};
\draw[->] (Ka)-|(suma2)
node [right,pos=0.93]{$+$};

\draw[->] (suma2) -- (sumv2)
node [right,pos=0.8]{$+$};
\draw[->] (sumv2) -- (sumx2)
node [right,pos=0.8]{$+$};
\draw[->] (sumx2) -- (in1)
node [above,pos=0.6]{$\vect{a}_c$};

\draw[->] (refin) -- (refin-|inner.west)
node [above,pos=0.295]{$\vect{j}_{ref}$, $\vect{s}_{ref}$, ${\psi}_{ref}$, ${\dot{\psi}}_{ref}$, ${\ddot{\psi}}_{ref}$};

\draw[->] (xrefin) -- (sumx)
node [above,pos=0.3]{$\vect{x}_{ref}$}
node [above,pos=0.9]{$+$};
\fill [fill=myblue,opacity=1.0] (vin-|sumx)++(-0.25em,-0.25em) rectangle ++(0.5em,0.5em);
\fill [fill=myblue,opacity=1.0] (ain-|sumx)++(-0.25em,-0.25em) rectangle ++(0.5em,0.5em);
\fill [fill=myblue,opacity=1.0] (ain-|sumv)++(-0.25em,-0.25em) rectangle ++(0.5em,0.5em);
\draw[->] (vin) -- (sumv)
node [above,pos=0.15]{$\vect{v}_{ref}$}
node [above,pos=0.92]{$+$};
\draw[->] (ain) -- (suma2)
node [above,pos=0.057]{$\vect{a}_{ref}$}
node [above,pos=0.97]{$+$};
\draw[->] (midwayain) -- (suma1)
node [left,pos=0.8]{$+$};

%\node [sum,right of =xrefin,node distance = 4 em](sumxb){};
%\node [gain,right of=sumx, node distance = 5em](Kxb){$\vect{K_x}$};
%\node [sum,below of = oldspotKx, node distance = 3em](sumvb){};
%\node [sum,below of = oldspotKv, node distance = 5.5em](suma1b){};
%\node [gain,right of=sumv, node distance = 5em](Kvb){$\vect{K_v}$};
%\node [gain,right of=suma1, node distance = 4em](Kab){$\vect{K_a}$};
%\node [sum](suma2b)at(ain-|sumv2){};
%\node [sum,right of=oldspotKv, node distance = 8em](sumv2b){};
%\node [sum,right of=oldspotKx, node distance = 12em](sumx2b){};
	
\end{tikzpicture}}
\caption{Position and Velocity Control. The blue area contains the PD control design as described in Section \ref{sec:trajtrackacc}.} \label{fig:diagram2}
\end{figure*}
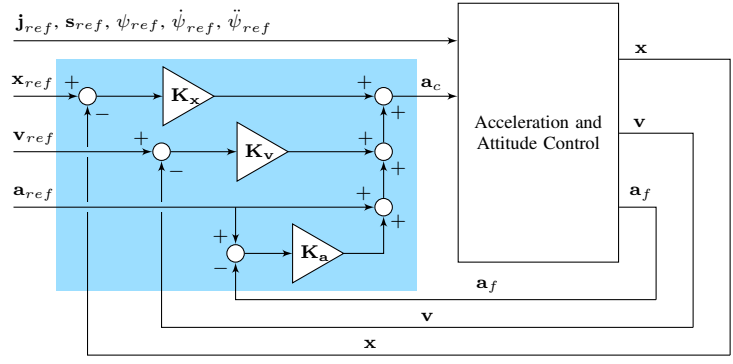
\begin{figure*}
\scriptsize
\centering
{\begin{tikzpicture}[auto, node distance=6em,>=latex']
		\fill [fill=myblue,opacity=1.0] (1.5em,-7em) rectangle ++(20em,9.5em);
	\fill [mygreen,opacity=1.0] (5em,-19em) rectangle ++(16.5em,11em);
%	\fill [mygreen,opacity=1.0] (20.5em,-22em) rectangle ++(7em,6em);
	\fill [myred,opacity=1.0] (23em,-13.8em) rectangle ++(22.5em,11.8em);
%	\fill [myred,opacity=1.0] (36em,-16em) rectangle ++(10.5em,-3em);
	
	\fill [myyellow,opacity=1.0] (47.5em,-9em) rectangle ++(13em,7em);
	
	\node [tmp] (acin) {};
	
	\node [sum,right of=acin, node distance = 3 em](suma){};
	\node [sum,right of=suma, node distance = 4 em](sumaa){};
	\node [block,right of=sumaa, node distance = 6.5 em](norm2){$-\|\cdot\|_2$};
	\node [gain,right of=norm2, node distance = 5.5 em](mgain){$m$};
	
	\node [tmp,right of=sumaa,node distance = 2em](tbcdown){};
	\node [block,below right=1.5em and -4.5em of norm2,text width=8 em](thrustdecomp){Attitude Increment Computation};
	\node [tmp,above left=-0.5em and 0em of thrustdecomp](attincr1in){};
	\node [tmp,below left=-0.5em and 0em of thrustdecomp](attincr2in){};
	
	\node [tmp,right of=thrustdecomp,node distance = 8 em](concat1up){};
	\node [tmp,below of=concat1up, node distance = 3 em](concat1down){};
	\node [tmp] (psirefin) at (acin |- concat1down){};
	\node [tmp,below right=-0.5em and 0em of concat1up](concat1){};
	
	\node [tmp] (belowtdec) at (thrustdecomp |- psirefin){};
	
	\node [block,right of=thrustdecomp,node distance = 13 em,text width=7 em](sumxi){Error Angle Computation};
	\node [gain,right of=sumxi,node distance = 8 em,inner sep=0.25em](kxi){$\vect{K_\xi}$};
	\node [sum,right of =kxi, node distance = 4.5 em](sumu){};
	\node [sum,right of =sumu, node distance = 3 em](sumu2){};
	\node [sum,right of =sumu2, node distance = 6 em](dOmegasum){};
	\node [gain,right of =dOmegasum, node distance = 4 em](J){$\vect{J}$};
	\node [sum,right of =J, node distance = 4.5 em](summu){};
	
	\node [tmp,below left=3.5em and -2. em of dOmegasum](Sinv){};
	\node [tmp,below left=3.5em and -2. em of Sinv](L){};
	\node [tmp,below right = 0.75 em and 0.75em of Sinv](nexttoSinv){};
	\node [tmp,right of= nexttoSinv,node distance =1.5em](nexttoSinv2){};
	\node [tmp,above right=-0.5em and 0em of L](L1in){};
	\node [tmp,below of = L1in,node distance = 2em](L2in){};
	\node [tmp](nexttoL2in) at (L1in -| nexttoSinv2){};
	\node [tmp,below of=nexttoSinv2,node distance = -5.8em](nexttoSinv3){};
	
	\node [tmp,below of=psirefin,node distance=2 em](jrefin){};
	\node [block,below right = -0.5em and 6em of jrefin](jgain){Jerk Tracking};
	\node [tmp,above left=-0.5em and 0em of jgain](jgain1in){};
	\node [tmp,below of = jgain1in,node distance = 2em](jgain2in){};
	\node [tmp,below right = 0em and  0 em of jgain](jin5){};
	\node [tmp](jinmid) at (jgain |-jin5){};
	\node [tmp,below right = 0em and -2 em of jgain](jin1){};
	\node [tmp,below left = 0em and -2 em of jgain](jin2){};
		
	\node [tmp,right of=jgain, node distance = 9 em](concat2up){};
	\node [tmp,below of=concat2up, node distance = 3 em](concat2down){};
	\node [tmp,below right=-0.5em and 0em of concat2up](concat2){};

	\node [tmp,below of=jrefin,node distance = 2 em](dpsirefin){};
	
	\node [tmp,right of=dpsirefin, node distance = 1.5em](halfwaydpsirefin){};
	
	\node [sum,right of=jgain,node distance=23em] (dxisum){};
	\node [biggain,below of=kxi, node distance=6 em](kdxi){$\vect{K_{\Omega}}$};

	\node [tmp,below of=dpsirefin,node distance=3 em](srefin){};
	\node [block,below right = -0.5em and 13em of srefin](sgain){Snap Tracking};
	\node [tmp,below right = 0em and  0 em of sgain](sin5){};
	\node [tmp](sinmid) at (sgain |-sin5){};
	\node [tmp,right of = sinmid, node distance = 3 em](sin4){};
	\node [tmp,right of = sinmid, node distance = 1 em](sin3){};
	\node [tmp,left of = sinmid, node distance = 1 em](sin2){};
	\node [tmp,left of = sinmid, node distance = 3 em](sin1){};
	
	\node [tmp,above left=-0.5em and 0em of sgain](sgain1in){};
	\node [tmp,below of = sgain1in,node distance = 2em](sgain2in){};
	\node [tmp,right of=sgain, node distance = 9 em](concat3up){};
	\node [tmp,below of=concat3up, node distance = 3 em](concat3down){};
	\node [tmp,below of=srefin,node distance = 2 em](ddpsirefin){};
	\node [tmp,right of=ddpsirefin, node distance = 1.5em](halfwayddpsirefin){};
	
	\node [bigblock2, below right = 5 em and -1 em of J](inner){Motor Control};
	\node [tmp,above left=-0.5em and 0em of inner](innerout1){};
	\node [tmp,above left=-2.5em and 0em of inner](innerout2){};
	\node [tmp,above left=-4.5em and 0em of inner](innerout3){};
	
	\node [tmp,above right=-0.5em and 0em of inner](innerin1){};
	\node [tmp,above right=-2.5em and 0em of inner](innerin2){};
	
	\node [tmp,below left=-0.5em and 0em of inner](inneroutl3){};
	\node [tmp,below left=-2.5em and 0em of inner](inneroutl2){};
	\node [tmp,below left=-4.5em and 0em of inner](inneroutl1){};
	\node [tmp,below left=-6em and 0em of inner](inneroutl){};
	
	\node [tmp,below right=-0.5em and 0em of inner](inneroutr3){};
	\node [tmp,below right=-2.5em and 0em of inner](inneroutr2){};
	\node [tmp,below right=-4.5em and 0em of inner](inneroutr1){};
	\node [tmp,right of = inneroutr3,node distance = 3 em](midafout){};
	
	\node [tmp,left of=innerout3,node distance = 9.5em](S){};
	\node [tmp,below right = 0.75 em and 0.75em of S](nexttoS){};
	\node [tmp](nexttoS2) at (nexttoS -| nexttoSinv2){};
	\node [tmp,right of = nexttoS2,node distance = 0em](newnexttoS2){};
	\node [tmp,left of=innerout3,node distance = 6.4em](nexttoS3){};
	
	\node [tmp,below left = 1 em and -1 em of jgain](bzgainmarker1){};
	\node [tmp,below of=inneroutl3,node distance = 5.5em](bzgainmarker2){};
	%\node [gainleft](bzgain) at (bzgainmarker1 |- bzgainmarker2){$\vect{b}_z$};

	\node [tmp, below of=bzgainmarker1,node distance = 12.63em](bzgain){};
		\node [tmp] (rightofbzgain) at (bzgain -| jin2){};
	\node [tmp,above right = 0.75 em and 0.75em of bzgain](nexttobzgain){};
	
	\draw [->] (inneroutl3) -| ++ (- 1em,-5.5em)
	 -| node [above,pos=0.19]{$(\tau\vect{b}_z)_f$} 	 
	 ++ (-48.5em,24.5em)  -| node [above,pos=0.18]{$(\tau\vect{b}_z)_f$} (sumaa)
	 node [right,pos=0.9]{$+$};

	\draw [->] (inneroutl2) -| ++ (- 2em,-5.5em) -|
	node [left,pos=0.94]{$\tau_f$} 
	node [above,pos=0.195]{$\tau_f$} (sin1); 
		\draw [->] (inneroutl2) -| ++ (- 2em,-5.5em) -|
	node [left,pos=0.955]{$\tau_f$} (jin2); 
	
	\draw [->] (inneroutl) -| ++ (- 4em,-5.5em) -|
node [left,pos=0.9]{$\dot \tau_f$} 
node [above,pos=0.155]{$\dot \tau_f$} (sin3); 
	
%%	\draw[->](nexttoL2in) -- (L1in);

	\node [tmp,right of=S,node distance = 1.5 em](helpS){};
	
	\draw [->] (midafout) -- ++ (0em,-7.5em) -| (suma)
	node [above,pos=0.25]{$\vect{a}_f$}
	node [left,pos=0.98]{$\vect{a}_f$}
	node [left,pos=0.995]{$-$};
			\draw [->] (innerout3) -| ++(-5em,-5em) -| (dxisum)
	node [left,pos=0.88,inner sep=0.em]{$\vect{\Omega}_f$}
	node [left,pos=0.95]{$-$};
	\fill [fill=white,opacity=1.0] (sgain-|dxisum)++(-0.25em,-0.25em) rectangle ++(0.5em,0.5em);
	\draw [->] (innerout3) -| ++(-5em,-5em) -| (sin4)
	node [above,pos=0.13]{$\vect{\Omega}_f$}
	node [left,pos=0.8,inner sep=0.em]{$\vect{\Omega}_f$};

	\node [tmp,below of=newnexttoS2,node distance = 7em](help1){};
	\fill [fill=white,opacity=1.0] (sin1)++(-0.25em,-6.375em) rectangle ++(0.5em,0.5em);
	\draw [->] (inneroutl1) -| ++ (- 3em,-5.5em) -| (sin2)
	node[above,pos=0.18]{$\vect{\xi}$}
	node[left,pos=0.93,inner sep=0.4em]{$\vect{\xi}$};
	\draw [->] (inneroutl1) -| ++ (- 3em,-5.5em) -| (jin1)
	node[left,pos=0.95,inner sep=0.4em]{$\vect{\xi}$};
	\node [tmp,below of=jin1, node distance = 1 em](below-jin1){};
	\draw[->] (below-jin1) -| (thrustdecomp)
	node [left,pos=0.9]{$\vect{\xi}$};

	\draw [->] (innerout1) -- ++(-1 em,0em) |- ++(0 em,3em) -| (summu)
	node [above,near start]{$\vect{\mu}_f$}
	node [right,pos=0.93]{$+$};
	\draw [->] (innerout2) -- ++(-2 em,0em) |- ++(0 em,5em) -| (dOmegasum)
		node [right,pos = 0.7]{$\vect{\dot \Omega}_f$}
		node [right,pos = 0.95]{$-$};
	\draw [->] (inneroutr1) -- ++(5 em,0em)
		node [above,midway]{$\vect{x}$};
	\draw [->] (inneroutr2) -- ++(5 em,0em)
		node [above,midway]{$\vect{v}$};
	\draw [->] (inneroutr3) -- ++(5 em,0em)
		node [above,midway]{$\vect{a}_f$};
	\draw [->] (summu) -- ++(4.5 em,0em) |- (innerin1)
		node [left,pos=0.3]{$\vect{\mu}_c$};
	\draw [->] (mgain) -- node [above,pos=0.1]{$T_c$}
	++(45 em,0em) |- (innerin2)
	node [above,near end]{$T_c$};
	
	\draw [->] (acin) -- (suma)
	node [above,pos=0.4]{$\vect{a}_c$}
	node [above,very near end]{$+$};
	
	\draw [->] (suma) -- (sumaa)
	node [above,very near end]{$+$};
	
	\draw [->] (sumaa) -- (norm2)
	node [above,pos = 0.42]{$(\tau \vect{b}_z)_c$};
	
	\draw [->] (norm2) -- (mgain)
	node [above,midway]{$\tau_c$};
	
	\draw [->] (tbcdown) |- (attincr1in);
	
	\draw [->] (thrustdecomp) -- (sumxi)
	node [above,midway]{$\vect{\xi}_c$};
	\node [tmp] at (psirefin|-attincr2in)(temptest){};
	\fill [fill=myblue,opacity=1.0] (temptest-|suma)++(-0.25em,-0.25em) rectangle ++(0.5em,0.5em);
	\fill [fill=myblue,opacity=1.0] (temptest-|suma)++(0.875em,-0.25em) rectangle ++(0.5em,0.5em);
	\draw [->] (psirefin|-attincr2in) -- (attincr2in)
	node [above,pos=0.1] {$\psi_{ref}$};
	\draw [->] (sumxi) -- (kxi)
	node [above,midway]{$\vect{\xi}_e$};
	\draw [->] (kxi) -- (sumu)
	node [above,pos=0.85]{$+$};
	\draw [->] (sumu) -- (sumu2)
	node [above,pos=0.9]{$+$};
	\draw [->] (sumu2) -- (dOmegasum)
	node [above,pos=0.39]{$\vect{\dot\Omega}_c$}
	node [above,pos=0.9] {$+$};
	\draw [->] (dOmegasum) -- (J);
	\draw [->] (J) -- (summu)
	node [above,pos=0.85]{$+$};

		\fill [fill=white,opacity=1.0] (jrefin-|suma)++(-0.25em,-0.25em) rectangle ++(0.5em,0.5em);
	\fill [fill=white,opacity=1.0] (jrefin-|suma)++(0.875em,-0.25em) rectangle ++(0.5em,0.5em);
	
			\fill [fill=white,opacity=1.0] (dpsirefin-|suma)++(-0.25em,-0.25em) rectangle ++(0.5em,0.5em);
	\fill [fill=white,opacity=1.0] (dpsirefin-|suma)++(0.875em,-0.25em) rectangle ++(0.5em,0.5em);
	
		\fill [fill=white,opacity=1.0] (srefin-|suma)++(-0.25em,-0.25em) rectangle ++(0.5em,0.5em);
\fill [fill=white,opacity=1.0] (srefin-|suma)++(0.875em,-0.25em) rectangle ++(0.5em,0.5em);

		\fill [fill=white,opacity=1.0] (ddpsirefin-|suma)++(-0.25em,-0.25em) rectangle ++(0.5em,0.5em);
\fill [fill=white,opacity=1.0] (ddpsirefin-|suma)++(0.875em,-0.25em) rectangle ++(0.5em,0.5em);

		\fill [fill=mygreen,opacity=1.0] (ddpsirefin-|jin1)++(-0.25em,-0.25em) rectangle ++(0.5em,0.5em);
\fill [fill=mygreen,opacity=1.0] (ddpsirefin-|jin2)++(-0.25em,-0.25em) rectangle ++(0.5em,0.5em);

		\fill [fill=mygreen,opacity=1.0] (srefin-|jin1)++(-0.25em,-0.25em) rectangle ++(0.5em,0.5em);
\fill [fill=mygreen,opacity=1.0] (srefin-|jin2)++(-0.25em,-0.25em) rectangle ++(0.5em,0.5em);
	
		\draw [->] (jrefin) -- (jgain1in)
	node [above,pos=0.17]{$\vect{j}_{ref}$};
	\draw [->] (dpsirefin) -- (jgain2in)
	node [above,pos=0.17]{$\dot \psi_{ref}$};
	\fill [fill=mygreen,opacity=1.0] (jgain-|thrustdecomp.south)++(-0.25em,-0.25em) rectangle ++(0.5em,0.5em);
	\draw [->] (jgain) -- (dxisum)
	node [above,pos=0.3]{$\vect{\Omega}_{ref}$}
	node [above,pos=0.95]{$+$};
	
	\draw [->] (srefin) -- (sgain1in)
	node [above,pos=0.08]{$\vect{s}_{ref}$};
	\draw [->] (ddpsirefin) -- (sgain2in)
	node [above,pos=0.08]{$\ddot \psi_{ref}$};

	\draw [->] (sgain) -| (sumu2)
	node [above,pos = 0.1,inner sep = 0.1em]{$\vect{\dot \Omega}_{ref}$}
	node [right,pos=0.985]{$+$};
		
	\draw [->] (dxisum) -- (kdxi);
	\draw [->] (kdxi) -| (sumu)
	node [right,pos=0.97]{$+$};

\end{tikzpicture}}
\caption{Acceleration and Attitude Control. The blue area contains the INDI linear acceleration and yaw control as described in Section \ref{sec:indilinacc}. The green area contains the computation of angular rate and angular acceleration references based on differential flatness as described in Section \ref{sec:diffflatness}. The red area contains the attitude and angular rate control as described in Section \ref{sec:trajtrack2}. The yellow area contains the INDI angular acceleration control as described in Section \ref{sec:attacccontrl}.} \label{fig:diagram1}
\end{figure*}
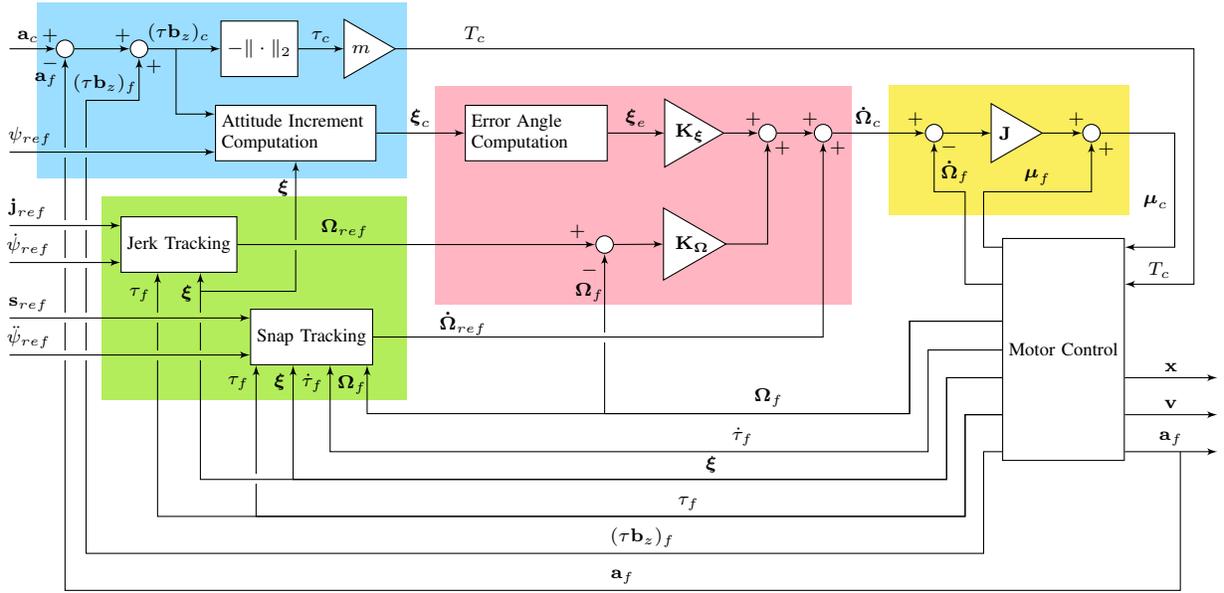
\begin{figure*}
\centering
\scriptsize
{\begin{tikzpicture}[auto, node distance=6em,>=latex']
	\fill [myblue,opacity=1] (1.7em,-2.5em) rectangle ++(12.5em,5em);
	\fill [mygreen,opacity=1] (15.2em,2.5em) rectangle ++(13.5em,-11em);
	\fill [myred,opacity=1] (35.5em,-18.5em) rectangle ++(23.5em,17.35em);
%	\fill [myyellow,opacity=1](45em,-1em) rectangle ++(10em,-4em);
%	\fill [myorange,opacity=1] (44em,-6.5em) rectangle ++(12em,-9.5em);
	
	\node [tmp] (left-of-numinv) {};
    
    \node [input, above of=left-of-numinv, node distance = 0.9em] (mu-c-input) {};
    
    \node [input, below of=left-of-numinv, node distance = 0.9em] (T-c-input) {};
    
    \node [block, right of=left-of-numinv, node distance=8em,text width=10em] (numinv) {Numerical Control Effectiveness Inversion};
    
    \node [lpfblock, right of=numinv, node distance = 14.1 em] (p) {$\mathpzc{p}(\cdot)$};
    
    \node [tmp, right of=numinv, node distance = 9em] (rightofnuminv){};
    
    \node [tmp, below of=rightofnuminv,node distance = 3.5 em](sumomega){};
    
    \node [sum, below of=sumomega,node distance = 2 em](newsumomega){};
    
    \node [tmp,right of=sumomega,node distance = 5.7 em,inner sep=0.15em](Komega){$\vect{K_\omega}$};
    
    \node [tmp,right of=sumomega,node distance = 1.3 em](rightofsumomega){};
    
    \node [lpfblock,right of = newsumomega, node distance = 3.5 em](Kint){$\int$};
    
    \node [gain,right of=Kint,node distance = 4 em,inner sep=0.15em](Kiomega){};
    
    \node [sum,right of = p,node distance = 5 em](newsumomega2){};
    \node [tmp,right of = Komega,node distance = 5 em](sumomega2){};
    
    \node [uavbigblock,right of = sumomega2,node distance = 5.5 em](UAV){UAV};
    
    \node [tmp,right of = UAV,node distance = 20 em](middlerightofUAV){};
    \node [tmp,right of = UAV,node distance = 6 em](leftmiddlerightofUAV){};
    \node [tmp,right of = middlerightofUAV,node distance = 7 em](rightofmiddlerightofUAV){};
    
    \node [tmp,above of = rightofmiddlerightofUAV,node distance = 5em](xout){};
    \node [tmp,above of = rightofmiddlerightofUAV,node distance = 3.8em](vout){};
    \node [tmp,above of = rightofmiddlerightofUAV,node distance = 2.6em](xiout){};
    \node [tmp,below of = rightofmiddlerightofUAV,node distance = -0.5em](aout){};
    \node [tmp,below of = rightofmiddlerightofUAV,node distance = 2.75em](Omegaout){};
    \node [tmp,below of = rightofmiddlerightofUAV,node distance = 7em](omegaout){};
    \node [tmp,left of = omegaout,node distance = 12em](lomegaout){};
    \node [tmp,right of = lomegaout,node distance = 3 em](sumG){};
    \node [tmp,right of = sumG,node distance = 2 em](dcat){};
    
    \node [tmp,above right = -0.5 em and 0 em of dcat](dcatout1){};
    \node [tmp,below right = -0.5 em and 0 em of dcat](dcatout2){};
    \node [tmp](mufout) at (dcatout1 -|rightofmiddlerightofUAV){};
    \node [tmp](taufout) at (dcatout2 -|rightofmiddlerightofUAV){};
    \node [tmp,right of = dcatout2,node distance = 2.8 em](mgain){$\frac{1}{m}$};

    \node [tmp,left of=xiout,node distance = 20.25 em](xihalfway){};
    \node [tmp,below of=xihalfway,node distance = 1 em](belowxihalfway){};
    \node [tmp,below left=2.3em and 2.3 em of belowxihalfway](leftbelowxihalfway){};
    \draw [-] (xihalfway) -- (belowxihalfway);
    \draw [->] (belowxihalfway) -- (leftbelowxihalfway);
    
    \node [tmp,left of=xiout,node distance = 11.5 em](xihalfway2){};
    \node [tmp,below of=xihalfway2,node distance = 7 em](belowxihalfway2){};
    \node [tmp,below left=2.75em and 2.75 em of belowxihalfway2](leftbelowxihalfway2){};
    \draw [-] (xihalfway2) -- (belowxihalfway2);
    \draw [->] (belowxihalfway2) -- (leftbelowxihalfway2);

    \node [gain,below of = leftmiddlerightofUAV,node distance = -0.5em](LPFa){$\vect{R}$};
    
    \node [sum,right of=LPFa,node distance = 4 em](R){};
    \node [tmp,above of=R,node distance= 1.5 em](giorig){};
    \node [lpfblock,right of=R,node distance = 9em](suma){LPF};
    
    \node [tmp,above of = Omegaout,node distance = 0.7em](Omegafout){};
    \node [tmp,below of = Omegaout,node distance = 0.7em](dotOmegafout){};
    \node [tmp,above of = lomegaout,node distance = 1.2em](omegafout){};
    \node [tmp,below of = lomegaout,node distance = 1.2em](dotomegafout){};
    \node [tmp,below of = dotomegafout,node distance = 3 em](dotprod){$\langle\cdot,\cdot\rangle$};
    \node [tmp,right of =dotprod,node distance = 5.5 em,inner sep=0.5,minimum width=3.5em](dottfgain){};
    \node [tmp] (dtaufout) at (dottfgain -| taufout){};
    
    \node [tmp,left of=omegafout,node distance = 12 em](midomegaout){};
    \node [lpfblock,below of = leftmiddlerightofUAV,node distance = 2.75em](LPFOmega){LPF};
    \node [lpfblock,right of = midomegaout,node distance = 3em](LPFomega){$\|\cdot\|^{2}_2$};
    \node [gain,right of=LPFomega, node distance = 4 em](inertial){};
    \draw [->](LPFomega) -- (inertial)
    node [right,pos=1,inner sep=-0.1em]{$\frac{-k_\tau}{m}$};
    \node[gain,right of=inertial,node distance = 4.5em](inertial2){};
    \draw [->](inertial) -- (inertial2)
    node [above,pos=0.5]{$\tau$}
    node [right,pos=1,inner sep=0.4em]{$\vect{b}_z$};
    \node [lpfblock,right of = inertial2,node distance = 4.5em](LPFinertial){LPF};
    \draw [->] (inertial2) -- (LPFinertial);
    \draw [->](LPFinertial) -- (LPFinertial -| aout)
    node [above,pos = 0.7]{$(\tau \vect{b}_z)_f$};
    \node [tmp] (temp-belowmidomegaout) at (midomegaout |- LPFomega){};
    \node [tmp,left of=temp-belowmidomegaout,node distance=0em](belowmidomegaout){};
    \node [tmp,left of=dotomegafout,node distance = 3.9 em](midomegafout){};
    \node [tmp,below left=-0.5em and 0 em of dotprod](dotinlow){};
    \node [tmp,above left=-0.5em and 0 em of dotprod](dotinhi){};
    \node [tmp](belowmidomegafout) at (midomegafout |- dotinlow){};
    \node [tmp](abovemidomegafout) at (midomegafout |- omegafout){};
%    \node [tmp,right of = abovemidomegafout, node distance = 1.5 em](middomegafout){};
   
   \node[lpfblock,below of=LPFOmega,node distance = 6.5 em](squared){$\cdot^{\circ 2}$};
   \node[lpfblock,right of=squared,node distance = 4.5 em](LPFmoment){LPF};
   \node[smallishgain,right of=LPFmoment,node distance = 6.5 em](G1){};
   
   \draw[->](midomegaout) |- (squared);
   \draw[->](squared) |- (LPFmoment);
   \draw[->](LPFmoment) |- (G1)
   node [above,pos=0.75,inner sep=0.1em]{$(\vect{\omega}^{\circ 2})_f$}
   node [below,pos=0.75,inner sep=0.1em]{$\dot{(\vect{\omega}^{\circ 2})_f}$}
   node [right,pos=1,inner sep=0.05em]{$\vect{G}_1$};
   
%   \node[lpfblock,below of=squared,node distance = 3 em](diff){$\frac{d}{dt}$};
   \node[lpfblock,below of=LPFmoment,node distance = 4 em](LPFdiff){LPF};
   \node[smallishgain,right of=LPFdiff,node distance = 6.5 em](G2){};
   
      \draw[->](midomegaout) |- (LPFdiff);
%   \draw[->](diff) |- (LPFdiff);
   \draw[->](LPFdiff) |- (G2)
   node [above,pos=0.75,inner sep=0.1em]{$\vect{\dot \omega}_f$}
   node [right,pos=1,inner sep=0.05em]{$\vect{G}_2$};
   
   \node[sum,below right = 1.05em and 2em of G1](sumG1G2){};
   \node[concat,right of=sumG1G2,node distance=2em](Tmmux){};
   	\node[tmp,below right = -0.25em and 0em of Tmmux](tmux2out){};
   \node[smallgain,right of=tmux2out,node distance=1.5em](divm){};
   
   \draw[->](G1) -| (sumG1G2)
	node [left,very near end]{$+$};
   \draw[->](G2) -| (sumG1G2)
	node [left,very near end]{$+$};
   \draw[->](sumG1G2) -- (Tmmux);
   
   \node[tmp,above right = -0.25em and 0em of Tmmux](tmux1out){};
   \draw[->](tmux1out)--(tmux1out -| aout)
   node[above,pos=0.7]{$\vect{\mu}_f$};

	\draw[->](tmux2out)--(divm)
	node [right,pos=1,inner sep=-0.05em]{$\frac{1}{m}$};
	\draw[->](divm)--(divm-| aout)
	node[above,pos=0.45]{$\tau_f$}
	node[below,pos=0.45]{$\dot \tau_f$};
   
%   \node[lpfblock,below of=LPFdiff,node distance = 3 em](prod){$\langle\cdot,\cdot \rangle$};
%   \node[lpfblock,right of=prod,node distance = 4.5 em](LPFder){LPF};
%   \node[gain,right of=LPFder,node distance = 3.8 em](dergain){};
%   
%   \node[tmp,above left= -0.5 em and 0em of prod](prodin1){};
%   \node[tmp,below left= -0.5 em and 0 em of prod](prodin2){};
%   \draw[->] (diff.east)++(0.5em,0em)|-(prodin1){};
%   \draw[->] (midomegaout)|-(prodin2){};
%   
%   \draw[->](prod)--(LPFder);
%   \draw[->](LPFder)--(dergain)
%   node [right,pos=1,inner sep=-0.15em]{$\frac{2k_\tau}{-m}$};

	\draw [->] (mu-c-input) -- (mu-c-input-|numinv.west)
	node [above,near start]{$\vect{\mu}_c$};
	\draw [->] (T-c-input) -- (T-c-input-|numinv.west)
	node [above,near start]{$T_c$};
	\draw [->] (numinv) -- (p)
	node [above,midway]{$\vect{\omega}_c$};
	\draw [->] (rightofnuminv) -- (newsumomega)
	node [left,very near end]{$+$};
	%\draw [->] (newsumomega) -- (Komega);
	\draw [->] (newsumomega) -- (Kint);

	\draw [->] (p) -- (newsumomega2)
	node [above,pos=0.8]{$+$};
	\draw [->] (Kiomega) -| (newsumomega2)
	node [left,pos=0.95]{$+$};
%	\draw [->] (Komega) -- (sumomega2)
%	node [above,very near end]{$+$};
	\draw [->] (newsumomega2) -- (UAV.west|-newsumomega2)
	node [above,pos=0.5]{$\vect{\zeta}$};
	
	\draw [->] (UAV.east|-xout) -- (xout)
	node [above,pos=0.93,inner sep = 1pt]{$\vect{x}$};
	\draw [->] (UAV.east|-vout) -- (vout)
	node [above,pos=0.93,inner sep = 1pt]{$\vect{v}$};
	\draw [->] (UAV.east|-xiout) -- (xiout)
	node [above,pos=0.93,inner sep = 1pt]{$\vect{\xi}$};
	\draw [->] (UAV.east|-aout) -- (LPFa)
	node [above,pos=0.45]{$\vect{a}^b$};
	\draw [->] (UAV.east|-Omegaout) -- (LPFOmega)
	node [above,midway]{$\vect{\Omega}$};
	\draw [-] (UAV.east|-midomegaout) -- (midomegaout)
	node [above, pos=1]{$\vect{\omega}$};
	\draw [->] (belowmidomegaout) -- (LPFomega);

	\draw [->](LPFa)--(R)	node [below,pos=0.9]{$+$};
	\fill [fill=myred,opacity=1.0] (suma-|xihalfway2)++(-0.25em,-0.25em) rectangle ++(0.5em,0.5em);
	\fill [fill=myred,opacity=1.0] (LPFOmega-|xihalfway2)++(-0.25em,-0.25em) rectangle ++(0.5em,0.5em);
	\draw [->](R) -- (suma);
	\draw [->] (giorig) -- (R)
	node [left,near start]{$g\vect{i}_z$}
	node [right,near end]{$+$};
	\draw [->] (suma) -- (aout)
	node [above,pos=0.73,inner sep = 1pt]{$\vect{a}_f$};
	
	\draw [->] (LPFOmega.east) -- (LPFOmega.east -|Omegafout)
	node [above,pos=0.91]{$\vect{\Omega}_f$, $\vect{\dot \Omega}_f$};
%	\draw [->] (LPFOmega.east|-dotOmegafout) -- (dotOmegafout)
%	node [above,pos=0.91,inner sep=0.2pt]{$\vect{\Omega}_f$};
%	\draw [->] (LPFomega.east|-omegafout) -- (omegafout)
%	node [above,pos=0.6,inner sep=0.2pt]{$\vect{\dot \omega}_f$}
%	node [right,at end,inner sep=0em]{$\vect{G}_2$};

	\draw [->] (midomegaout.south) -- ++(0,-3.45em) -| (newsumomega.south)
	node [above,near start]{$\vect{\omega}$}
	node [left,pos=0.95]{$-$};
	
		\draw [->] (midomegaout.south) -- ++(0,-3.45em) -| (numinv.south)
	node [right,pos=0.9]{$\vect{\omega}$};
	
%	\draw [->] (omegafout) -| (sumG.north)
%	node [right,near end] {$+$};
%	\draw [->] (dotomegafout) -| (sumG.south)
%	node [right,near end] {$+$};
%	\draw [->] (sumG) -- (dcat);
%	
%	\draw [->] (belowmidomegafout) -- (dotinlow);
%	\draw [-] (middomegafout) -- ++(0em,-0.3em);
%	\draw [->] (middomegafout)+(0em,-0.3em) |- (dotinhi);

%	
%    \draw [->] (dcatout1) -- (mufout)
%    node [above,pos=0.8]{$\vect{\mu}_f$};
%    \draw [->] (dcatout2) -- (mgain)
%    node [above,midway]{$T_f$};
%    \draw [->] (mgain) -- (taufout)
%    node [above,pos=0.43]{$\tau_f$};
%    \draw [->] (dottfgain) -- (dtaufout)
%    node [above,pos=0.7]{$\dot \tau_f$};

%	\draw [->] (LPFomega.east|-dotomegafout) -- (dotomegafout)
%	node [right,at end,inner sep=0em]{$\vect{G}_1$}
%	node [above,pos = 0.25,inner sep=0.2pt]{$\vect{\omega}_f$};
%	
%	\node [tmp](crossleft) at (LPFomega.east|-dotomegafout){};
%	\node [tmp, right of = crossleft, node distance = 1.5em](crossleft2){};
%	\node [tmp, right of = crossleft2, node distance = 1em](crossright){};
%	\draw [crossline,-] (crossleft2) -- (crossright);
		\draw [->] (Kint) -- (Kiomega)
	node [right,at end,inner sep=0em]{$\vect{K_{I_\omega}}$};

\end{tikzpicture}}
\caption{Motor Control and Computation of Filtered Signals. The blue and green areas contain the moment and thrust control (including motor speed command saturation resolution), and the motor speed control, respectively. Both are described in Section \ref{sec:motorspeedctrl}. The \textit{UAV} block represents the UAV hardware, including ESCs, motors, and sensors. The red area contains the computation of filtered signals based on IMU and optical encoder measurements.} \label{fig:diagram3}
\end{figure*}
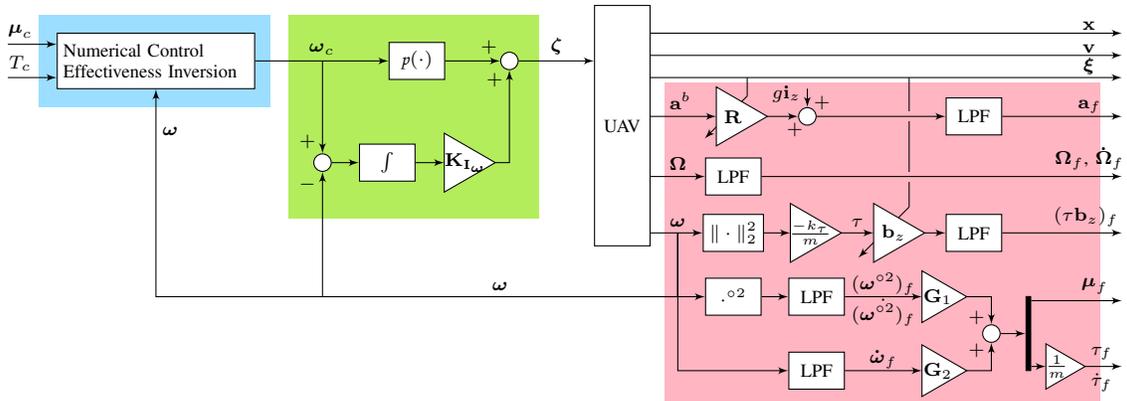

The controller utilizes a vehicle state estimate consisting of position, velocity, and attitude.
Additionally, motor speed measurements are obtained from optical encoders, and
linear acceleration and angular rate measurements are obtained from the inertial measurement unit (IMU).
For the application of incremental angular acceleration control, angular acceleration measurements are obtained by numerical differentiation of the measured angular rate.
A low-pass filter (LPF) is required to alleviate the effects of noise, \eg, airframe vibrations, on measurements obtained directly from the IMU.
We denote the LPF outputs using the subscript $f$, \eg, by $\vect{\Omega}_f$ and $\vect{\dot \Omega}_f$ for the angular rate output and its derivative, respectively.
The gravity-corrected LPF acceleration output in the inertial reference frame is obtained as follows:
\begin{equation}\label{eq:gravcorrection}
\vect{a}_f = (\vect{R}\vect{a}^b +g\vect{i}_z )_f.
\end{equation}

\subsection{PD Position and Velocity Control}\label{sec:trajtrackacc}
Position and velocity control is based on two cascaded proportional-derivative (PD) controllers. The resulting controller is mathematically equivalent to the following single expression:
\begin{multline}\label{eq:pd}
\vect{a}_c = \vect{K}_{\vect{x}} \left(\vect{x}_{ref}-\vect{x}\right) + \vect{K}_{\vect{v}} \left(\vect{v}_{ref}-\vect{v}\right)\\+ \vect{K}_{\vect{a}} \left(\vect{a}_{ref}-\vect{a}_f\right) + \vect{a}_{ref}
\end{multline}
with $\vect{K_{\bullet}}$ indicating diagonal gain matrices.
The subscript $ref$ is used to indicate values obtained directly from the reference trajectory. In contrast, the subscript $c$ indicates commanded values that are computed in one of the control loops. For example, $\vect{a}_{ref}$ is obtained directly from the reference trajectory as the second derivative of $\vect{x}_{ref}$, while $\vect{a}_c$ is computed based on \eqref{eq:pd} and includes terms based on the position, velocity, and acceleration deviations.
The first three terms in \eqref{eq:pd} ensure tracking of position and velocity references, while the final term serves as a feedforward input to ensure tracking of the reference acceleration.
The control utilizes the inertial reference frame with --- in our implementation --- identical gains for the horizontal $\vect{i}_x$- and $\vect{i}_y$-directions,
but separately tuned gains for the vertical $\vect{i}_z$-direction.
The commanded acceleration is used to calculate thrust and attitude commands, as will be shown in the next section.

\subsection{INDI Linear Acceleration and Yaw Control}\label{sec:indilinacc}
Existing literature presents the derivation of an INDI linear acceleration controller using  Taylor series approximation~\cite{smeur2017cascaded}. In this section, we arrive at equivalent control equations through an intuitive derivation that follows the practical working of the INDI notion based on estimation of the external force acting on the quadrotor.

An expression for the external force in terms of measured acceleration and specific thrust is obtained by rewriting \eqref{eq:vdot}, as follows:
\begin{equation}
\vect{f}_{ext} = m \left(\vect{a}_f - (\tau \vect{b}_{z})_f - g\vect{i}_z\right),\label{eq:fext}
\end{equation}
where $\tau$ is the specific thrust calculated according to \eqref{eq:mutau} using motor speed measurements.
Identical LPFs must be used to ensure that equal phase lag is incurred by acceleration and thrust measurements~\cite{smeur2015adaptive}.
Note that the specific thrust vector and the linear acceleration (cf. \eqref{eq:gravcorrection}) are both transformed to the inertial reference frame prior to filtering.
This order is appropriate because the external force in the inertial reference frame $\vect{f}_{ext}$ is assumed to be slow-changing relative to the LPF dynamics, as described in Section \ref{sec:diffflatness}.
Substitution of \eqref{eq:fext} into \eqref{eq:vdot} gives the following expression for the current acceleration:
\begin{align}
\vect{a} &= \tau \vect{b}_z +g\vect{i}_z + m^{-1}\vect{f}_{ext}\notag\\
&= \tau \vect{b}_z +g\vect{i}_z + m^{-1}\left(m \left(\vect{a}_f - (\tau \vect{b}_{z})_f - g\vect{i}_z\right)\right)\label{eq:accindi}\\
&=\tau \vect{b}_z- (\tau \vect{b}_{z})_f + \vect{a}_f.\notag
\end{align}
The specific thrust vector command that results in the commanded acceleration prescribed by \eqref{eq:pd} can be computed using the following incremental relation based on \eqref{eq:accindi}:
\begin{equation}\label{eq:accindi2}
(\tau \vect{b}_{z})_c = (\tau \vect{b}_{z})_f + \vect{a}_c - \vect{a}_f.
\end{equation}
The incremental nature of \eqref{eq:accindi2} enables the controller to achieve the commanded acceleration despite possible disturbances or modeling errors. If the commanded value is not obtained immediately, the thrust and attitude commands will be incremented further in subsequent control updates. This principle eliminates the need for integral action anywhere in the control design.

The thrust magnitude command is obtained as
\begin{equation}
T_c = -m\|(\tau \vect{b}_{z})_c\|_2\label{eq:deltatauc}
\end{equation}
with the negative sign following from the definition that thrust is positive in $\vect{b}_z$-direction. 
The incremental attitude command $\vect{\xi_{c}}$ represents the rotation from the current attitude to the commanded attitude and is obtained in two steps: first, the minimum rotation to align $-\vect{b}_z$ with the thrust vector command $(\tau \vect{b}_{z})_c$ is obtained; second, a rotation around $\vect{b}_z$ is added to satisfy the yaw reference $\psi_{ref}$.
For the first step, we transform the normalized thrust vector command to the current body-fixed reference frame, as follows:
\begin{equation}\label{key}
(-\vect{b}_{z})^b_c = \vect{\xi}^{-1} \circ(-\vect{b}_{z})_c \circ\vect{\xi}.
\end{equation}
The appropriate rotation to align the current $-\vect{b}_z$ with $(\tau \vect{b}_{z})_c$ is then given by
\begin{equation}\label{eq:attcmd_intermediate}
\vect{\bar \xi_{c}} = \widehat{\left[\begin{array}{c}
	1-\vect{i}_z^T (-\vect{b}_{z})^b_c\\
	-\vect{i}_z \times (-\vect{b}_{z})^b_c
	\end{array}\right]},
\end{equation}
where hat refers to quaternion normalization, \ie, $\hat{\vect{\xi}} = \nicefrac{\vect{\xi}}{\|\vect{\xi}\|_2}$.
For the second step, the yaw reference normal vector is first transformed to the intermediate attitude command frame, as follows:
\begin{multline}\label{key}
\vect{\bar n}_{\psi_{ref}} = (\vect{\xi}\circ\vect{\bar \xi_{c}})^{-1}\\\circ\left[\begin{array}{ccc}
\sin \psi_{ref} & -\cos \psi_{ref} & 0
\end{array}\right]^T\circ(\vect{\xi}\circ\vect{\bar \xi_{c}}).
\end{multline}
Next, we obtain the following rotation that makes $\vect{b}_x$ coincide with the plane defined by normal vector $\vect{\bar n}_{\psi_{ref}}$:
\begin{equation}\label{eq:attyawcmd}
\vect{\xi}_\psi = %\frac{1}{\sqrt{1 + \left(\frac{\bar n^1_{\psi_{ref}}}{\bar n^2_{\psi_{ref}}}\right)^2}}
\widehat{\left[\begin{array}{cccc}
		1&0&0&-\frac{\bar n^1_{\psi_{ref}}}{\bar n^2_{\psi_{ref}}}
			\end{array}\right]}^T.
\end{equation}
Equation \eqref{eq:attyawcmd} implicitly selects between tracking of $\psi_{ref}$ and $\psi_{ref} + \pi$ \si{rad} based on minimizing the magnitude of rotation.
Due to continuity of $\psi_{ref}$ this does not cause any unwanted switching, but it does prevent unwanted discontinuities such as a $\pi$ \si{rad} rotation around $\vect{b}_z$ to maintain yaw tracking when pitching through $\pm \nicefrac{\pi}{2}$ \si{rad}.
Note that \eqref{eq:attcmd_intermediate} and \eqref{eq:attyawcmd} incur singularities if $\vect{i}_{z} = (-\vect{b}_{z})^b_c$ and $\bar n^2_{\psi_{ref}} = 0$, respectively.
However, by computing the attitude command relative to the current attitude we move these singularities far away from the nominal trajectory.
Moreover, they are straightforwardly detected and resolved by selecting any direction of rotation.
Finally, the incremental attitude command is obtained as
\begin{equation}\label{key}
\vect{\xi}_c = \vect{\bar \xi}_c \circ \vect{\xi}_{\psi}.
\end{equation}

\subsection{PD Attitude and Angular Rate Control}\label{sec:trajtrack2}
In this section, we describe the attitude and angular rate controller.
This controller specifies the angular rate command and is thus solely based on angular kinematics. This has two major advantages compared to incorporating control torque or motor speeds. Firstly, the attitude controller does not take into account any model-specific parameters, such as the vehicle inertia matrix $\vect{J}$. Therefore the control design avoids discrepancies due to model mismatches and has vehicle-independent gains. Secondly, accurate torque control cognizant of the external moment $\vect{\mu}_{ext}$ can be performed separately using sensor-based INDI, as described in Section \ref{sec:attacccontrl}. This eliminates the need to incorporate a complicated disturbance model in the attitude controller, which further improves controller robustness and simplicity.

The three-element angle vector $\vect{\xi}_e$ associated with the incremental attitude command $\vect{\xi}_c$ is computed as follows:
\begin{equation}\label{eq:errorangle}
\vect{\xi}_e = \frac{2 \arccos\xi_c^w}{\sqrt{1- \xi_c^w\xi_c^w}}\left[\begin{array}{ccc}
\xi_c^x&\xi_c^y&\xi_c^z
\end{array}\right]^T.
\end{equation}
Using these error angles, the angular acceleration command is obtained as
\begin{equation}\label{eq:PD}
\vect{\dot \Omega}_c = \vect{K_\xi}\vect{\xi}_e + \vect{K_{\Omega}}\left( \vect{\Omega}_{ref}-\vect{\Omega}_f\right) + \vect{\dot \Omega}_{ref},
\end{equation}
where $\vect{\Omega}_{ref}$ and $\vect{\dot \Omega}_{ref}$ are the angular velocity and angular acceleration feedforward terms defined in \eqref{eq:angrateref} and \eqref{eq:angaccref}, respectively.
The resulting attitude controller not only tracks the attitude command, but also angular rate and acceleration. This enables tracking of trajectory jerk and snap, which is essential for accurate tracking of aggressive trajectories, as will be shown analytically in Section \ref{sec:analysis} and experimentally in Section \ref{sec:experiments}. In contrast, trajectory tracking control based on body rate inputs, \eg, using an off-the-shelf flight controller, is incapable of truly considering reference snap, because snap corresponds to the vehicle angular acceleration, as shown in \eqref{eq:angaccref}.

\subsection{INDI Angular Acceleration Control}\label{sec:attacccontrl}
Robust tracking of the angular acceleration command $\vect{\dot \Omega}_c$ is achieved through INDI control. 
We rewrite \eqref{eq:Omegadot} into the following expression for the external moment based on the measured angular rate, angular acceleration, and control moment:
\begin{equation}\label{eq:mu_ext}
\vect{\mu}_{ext} = \vect{J}\vect{\dot \Omega}_f-\vect{\mu}_f+\vect{\Omega}_f\times\vect{J}\vect{\Omega}_f
\end{equation}
with $\vect{\mu}_f$ the control moment in the body-fixed reference frame, obtained from the measured motor speeds by \eqref{eq:mutau} and low-pass filtering.
Analogous to the external force in Section \ref{sec:indilinacc}, the external moment $\vect{\mu}_{ext}$ is assumed slow-changing with regard to the LPF dynamics. Substitution of \eqref{eq:mu_ext} into \eqref{eq:Omegadot} then gives:
\begin{align}
\vect{\dot \Omega} &= \vect{J}^{-1}(\vect{\mu}+\vect{\mu}_{ext}-\vect{\Omega} \times \vect{J}\vect{\Omega})\notag\\
&= \vect{J}^{-1}(\vect{\mu}+( \vect{J}\vect{\dot \Omega}_f-\vect{\mu}_f+\vect{\Omega}_f\times\vect{J}\vect{\Omega}_f)-\vect{\Omega} \times \vect{J}\vect{\Omega})\notag\\
&= \vect{\dot \Omega}_f+\vect{J}^{-1}(\vect{\mu}-\vect{\mu}_f).\label{eq:omegadotmuf}
\end{align}
In \eqref{eq:omegadotmuf}, it is assumed that the difference between the gyroscopic angular momentum term and its filtered counterpart is sufficiently small to be neglected,
because the term is relatively slow changing compared to the angular acceleration and control moment, and moreover is second-order.
By inversion of the final line, we obtain the following incremental expression for the commanded control moment:
\begin{equation}\label{eq:mu_c}
\vect{\mu}_c = \vect{\mu}_f + \vect{J}\left(\vect{\dot\Omega}_c-\vect{\dot \Omega}_f\right).
\end{equation}

\subsection{Inversion-Based Moment and Thrust Control, and Integrative Motor Speed Control}\label{sec:motorspeedctrl}
In Section \ref{sec:indilinacc} and Section \ref{sec:attacccontrl}, we have found expressions for the commanded thrust $T_c$ and control moment $\vect{\mu}_c$, respectively.
Tracking of these commands requires control of the motor speeds, as evidenced by the direct relation given in \eqref{eq:mutau}.
State-of-the-art INDI implementations for quadrotors are based on linearization of this relation and do not accurately model transient behavior~\cite{smeur2015adaptive,smeur2017cascaded}.
Our proposed implementation is based on a nonlinear inversion of the control effectiveness and explicitly incorporates the motor response time constant, as such it provides a more accurate computation of control inputs.

\begin{figure}
	\centering
	\includegraphics[width=0.75\linewidth]{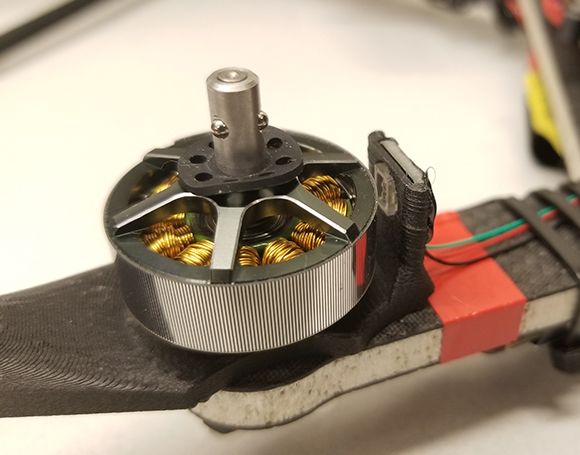}
	\caption{Motor (propeller removed) with optical encoder for rotational speed measurement. Note the optical encoder lens on the right, and the accompanying reflective strip on the motor hub.}\label{fig:opto}
\end{figure}

In order to achieve fast and accurate closed-loop motor speed control, we employ optical encoders that measure the motor speeds. The availability of motor speed measurements furthermore enables accurate calculation of the thrust and control moment, as required by the INDI controller in \eqref{eq:accindi2} and \eqref{eq:mu_c}.
In practice, the optical encoder, shown in Fig. \ref{fig:opto}, measures the motor rotational speed by detecting the passage of stripes on a reflective strip attached to the motor hub. As such, the optical encoder provides a high-rate, accurate, lightweight, and unintrusive manner to obtain the motor speed.

The motor speed corresponding to the commanded thrust and control moment is found by inverting the nonlinear control effectiveness equation \eqref{eq:mutau}. In order to do so, we estimate the effect of the motor speed command on the motor speed derivative using the following first-order model:
\begin{equation}\label{eq:motordyn}
\vect{\dot \omega} = \uptau_m^{-1} ({\vect{\omega}_c}-{\vect{\omega}})
\end{equation}
with $\uptau_m$ the motor dynamics time constant.
After equating to the control moment and thrust commands, the resulting equation,
\begin{equation}\label{eq:mutauc}
\left[\begin{array}{c}
\vect{\mu}_c\\
T_c
\end{array}\right]=\vect{G}_1 \vect{\omega}_c^{\circ 2} + \uptau_m^{-1}\vect{G}_2 ({\vect{\omega}_c}-{\vect{\omega}}),
\end{equation}
can be solved numerically, \eg, using Newton's method. Inversion of this nonlinear control effectiveness relation improves the accuracy of thrust and control moment tracking, when compared to the linearized inversion that does not consider the motor transient response as given by \eqref{eq:linearizedINDI}.

Inversion of \eqref{eq:mutauc} may lead to infeasible, \ie, saturated, motor speed commands.
We address this first by altering the control moment around $\vect{b}_z$.
Since the control effectiveness is relatively much smaller around this axis,
this is most likely to resolve the command saturation.
Moreover, it typically least affects vehicle stability and position tracking, since rotation purely around the $\vect{b}_z$-axis does not alter the thrust vector.
Let $\underaccent{\bar}\omega$ and $\bar \omega$ be respectively the minimum and maximum feasible motor speeds, then the set of $\vect{b}_z$ control momenta --- excluding $J_{r_z}$ contributions --- that result in feasible motor speed commands is
\begin{multline}\label{eq:feasmz}
\max \left\{\frac{k_{\mu_z}}{k_\tau}\left(4k_\tau \underaccent{\bar}\omega^2 + T_c \pm \left(\frac{\mu_c^y}{l_x} - \frac{\mu_c^x}{l_y}\right)\right),\right. \\
\left. -\frac{k_{\mu_z}}{k_\tau}\left(4k_\tau \bar\omega^2 + T_c \pm \left(\frac{\mu_c^y}{l_x} + \frac{\mu_c^x}{l_y}\right)\right)\right\} \leq \mu_c^z\\
 \leq \min \left\{\frac{k_{\mu_z}}{k_\tau}\left(4k_\tau \bar\omega^2 + T_c \pm \left(\frac{\mu_c^y}{l_x} - \frac{\mu_c^x}{l_y}\right)\right),\right. \\
\left. -\frac{k_{\mu_z}}{k_\tau}\left(4k_\tau \underaccent{\bar}\omega^2 + T_c \pm \left(\frac{\mu_c^y}{l_x} + \frac{\mu_c^x}{l_y}\right)\right)\right\}.
\end{multline}
If this set is non-empty, we set $\mu_c^z$ to equal the boundary closest to the original moment command.
The motor speed command is then obtained as
\begin{equation}\label{eq:simplemotorspeeds}
\vect{\omega}_c = \left(\vect{G}_1^{-1}\left[\begin{array}{c}
\vect{\mu}_c \\ T_c
\end{array}\right]\right)^{\circ \frac{1}{2}}.
\end{equation}
Note that due to $J_{r_z}$ contributions the actual $\vect{b}_z$ control moment will not exactly be equal to $\mu_c^z$.
However, we still obtain the feasible control moment that is closest to the original commanded moment, because $k_{\mu_z}$ and $J_{r_z}$ have identical signs in \eqref{eq:G1} and \eqref{eq:G2}, respectively.
If there exists no $\mu_c^z$ that results in feasible motor commands, we consider a reduction or limited increase in the thrust magnitude command $T_c$ based on the reasoning that application of thrust is only effective in the correct direction, \ie, at the correct vehicle pitch and roll.
Since adjustment of $T_c$ results in equal magnitude shift of the constraints, it is straightforward to verify whether there exists an acceptable value of $T_c$ such that the lower and upper boundaries in \eqref{eq:feasmz} coincide.
If so, $T_c$ is set to this value and $\mu_c^z$ to the feasible point, after which \eqref{eq:simplemotorspeeds} is used to compute the motor speed commands.
If not, $\mu_c^z$ is set to the average of the lower and upper boundaries in \eqref{eq:feasmz}, and any infeasible motor speed commands resulting from \eqref{eq:simplemotorspeeds} are clipped.

Finally, the throttle vector $\vect{\zeta}$ that contains the motor electronic speed control (ESC) commands is obtained as follows:
\begin{equation}\label{eq:motorcommander}
\vect{\zeta} = \mathpzc{p}(\vect{\omega_c}) +\vect{K_{I_\omega}}\int \vect{\omega}_c-\vect{\omega} \diff t
\end{equation}
with $\mathpzc{p}$ a vector-valued polynomial function relating motor speeds to throttle inputs. This function was obtained by regression analysis of static test data. Integral action is added to account for changes in this relation due to decreasing battery voltage. The measured motor speed signal $\vect{\omega}$ remains unfiltered here to minimize phase lag.

\section{Response Analysis}\label{sec:analysis}
Incremental control, and the tracking of high-order reference derivatives are two key aspects of our control design.
In this section, we theoretically verify the advantages of these features.
Namely, the improved robustness of incremental control in comparison to non-incremental control, and the improved trajectory tracking accuracy due to the consideration of high-order reference trajectory derivatives, \ie, jerk and snap.
The purpose of this section is to provide an intuitive understanding of how these aspects improve tracking performance.
In order to analyze the behavior of the closed-loop system, we use linearized dynamics and control equations, as the resulting simplifications allow for easier qualitative interpretation.
However, the observations in this section also apply to the full, nonlinear dynamics and control equations.
Our findings are validated and quantitatively assessed using real-life flights in Section \ref{sec:experiments}.

\begin{figure*}
\centering
\scriptsize
{\begin{tikzpicture}[auto, node distance=6em,>=latex']
\node [tmp](axrefin){};
\fill [myblue,opacity=1] (21em,-7.5em) rectangle ++(28.25em,12.5em);
\node [midgain,right of=axrefin,node distance = 3 em](again){};
\node [midgain,below of=again,node distance = 6 em](afgain){};
\node [sum,right of=afgain,node distance = 3 em](asum){};
\node [tmp] (adown) at (again-|asum){};
\node [sum,right of=asum,node distance = 3 em](thetasum){};
\node [midgain,right of=thetasum,node distance = 6.5 em](kxi){};
\node [smallblock,right of = adown, node distance = 3 em](a2j){$s$};
\node [tmp,right of = a2j, node distance = 3 em](jdown){};
\node [sum,below of = jdown, node distance =3 em](jsum){};
\node [midgain,right of =jsum,node distance = 3.5 em](kdxi){};
\node [sum,right of =kdxi,node distance = 4 em](jsum2){};
\node [smallblock,right of = jdown, node distance = 3.5 em](j2s){$s$};
\node [sum] (ssum) at (j2s-|jsum2){};

\node [sum,right of = ssum,node distance = 3 em](alphasum){};
\node [midpgain,right of =alphasum,node distance = 3 em](JyykG){};
\node [sum,right of = JyykG,node distance = 4 em](omegasum){};
\node [smallblock,right of = omegasum,node distance = 4 em](M){$M(s)$};
\node [tmp,right of = M,node distance = 3 em](omegadown){};
\node [midgain,right of = omegadown,node distance = 2 em](kg){};
\node [sum,right of = kg,node distance = 4 em](musum){};
\node [tmp,above of = kg,node distance = 3 em](muin){};
\node [midpgain,right of = musum,node distance = 3 em](Jyy1){};
\node [tmp,right of = Jyy1,node distance = 3 em](alphadown){};

\node [smallblock,right of = alphadown, node distance = 2.5em] (alpha2q){$\frac{1}{s}$};
\node [tmp,right of = alpha2q,node distance = 2.5 em](qdown){};
\node [smallblock,right of = qdown, node distance = 2.5em] (q2theta){$\frac{1}{s}$};
\node [tmp,right of = q2theta,node distance = 2.5 em](thetadown){};
\node [tmp,above of = thetadown,node distance = 3 em](finold){};
\node [tmp,left of = finold,node distance = 1.5 em](fin){};
\node [midgain,right of = thetadown,node distance = 3 em](theta2a){};
\node [midgain,right of = fin,node distance = 4.5 em](fgain){};
\node [sum,right of =theta2a,node distance = 3 em](fsum){};
\node [tmp,right of =fsum,node distance = 2.5 em](adown2){};
\node [tmp,right of =adown2,node distance = 2.5 em](aout){};

\node [smallblock,below of = M,node distance = 3 em](H1){$H(s)$};
\node [smallblock,below of = H1,node distance = 3 em](H2){$H(s)$};
\node [smallblock,below of = H2,node distance = 3 em](H3){$H(s)$};
\node [tmp,below of = H3,node distance = 2 em](nonH){};
\node [tmp](rightofnonH) at (nonH-|thetadown){};
\node [smallblock,below of = H3,node distance = 4 em](H4){$H(s)$};
\node [smallblock,below of = H4,node distance = 3 em](H5){$H(s)$};

\draw[->] (axrefin)--(again)
node[above,pos=0.1]{$a_{x,ref}$}
node[right,pos=1,inner sep = 0.1 em]{-$\frac{1}{g}$};
\draw[->] (again)--(a2j);
\draw[->] (a2j)--(j2s)
node[above,pos=0.5]{$q_{ref}$};
\draw[->] (j2s)--(ssum)
node[below,pos=0.8]{$+$}
node[above,pos=0.55]{$\alpha_{ref}$};
\draw[->] (ssum)--(alphasum)
node[below,pos=0.8]{$+$}
node[above,pos=0.5]{$\alpha_c$};
\draw[->] (alphasum)--(JyykG)
node[right,pos=1,inner sep = -0.075 em]{$\frac{J_{yy}}{k_G}$};
\draw[->] (JyykG)--(omegasum)
node[above,pos=0.65]{$+$};
\draw[->] (omegasum)--(M)
node[above,pos=0.5]{$\omega_c$};
\draw[->] (M)--(kg)
node[right,pos=1,inner sep = 0.1 em]{$k_G$};
\draw[->] (kg)--(musum)
node[above,pos=0.5]{$\mu_y$}
node[below,pos=0.7]{$+$};
\draw[->] (musum)--(Jyy1)
node[right,pos=1,inner sep = -0.075 em]{$\frac{1}{J_{yy}}$};
\draw[->] (Jyy1)--(alpha2q)
node[above,pos=0.5]{$\alpha$};
\draw[->] (alpha2q)--(q2theta)
node[above,pos=0.5]{$q$};
\draw[->] (q2theta)--(theta2a)
node[above,pos=0.5]{$\theta$}
node[right,pos=1,inner sep = 0.1 em]{-$g$};
\draw[->] (theta2a)--(fsum)
node[below,pos=0.5]{$+$};
\draw[->] (fsum)--(aout)
node[above,pos=0.5]{$a_x$};

\draw[->] (omegadown) |- (H1)
node[above,pos=0.0]{$\omega$};
\draw[->] (alphadown) |- (H2);
\draw[->] (qdown) |- (H3);
\draw[->] (thetadown) |- (H4);
\draw[->] (adown2) |- (H5);

\draw[->] (H1) -| (omegasum)
node[right,pos=0.9]{$+$}
node[left,pos=0.6]{$\omega_{f}$};
\draw[->] (H2) -| (alphasum)
node[right,pos=0.95]{$-$}
node[above,pos=0.3]{$\alpha_{f}$};
\draw[->] (H3) -| (jsum)
node[left,pos=0.96]{$-$}
node[above,pos=0.22]{$q_{f}$};
\draw[->] (rightofnonH)-|(thetasum)
node[right,pos=0.953]{$-$}
node[above,pos=0.383]{$\theta$};
\draw[->] (H4) -| (asum)
node[right,pos=0.965]{$+$}
node[above,pos=0.25]{$\theta_{f}$};
\node [tmp](leftofafgain) at (afgain -| axrefin){};
\draw[->] (H5) -| (leftofafgain) node[above,pos=0.25]{$(a_x)_{f}$}
-- (afgain)
node[right,pos=1,inner sep = 0.1 em]{$\frac{1}{g}$};

\draw[->] (afgain) -- (asum)
node[below,pos=0.5]{$+$};
\draw[->] (adown) -- (asum)
node[left,pos=0.9]{$+$};
\draw[->] (asum) -- (thetasum)
node[below,pos=0.8]{$+$};
\node[tmp,left of = kxi,node distance = 1 em](leftofkxi){};
\draw[crossline,-] (thetasum) -- (leftofkxi)
node[above,pos=0.2]{$\theta_e$};
\draw[->] (leftofkxi) -- (kxi)
node[right,pos=1,inner sep = 0.1 em]{$k_{{\theta}}$};
\draw[->] (kxi) -| (jsum2)
node[right,pos=0.9]{$+$};
\draw[->] (jdown) -- (jsum)
node[left,pos=0.8]{$+$};
\draw[->] (jsum) -- (kdxi)
node[right,pos=1,inner sep = 0.1 em]{$k_q$};
\draw[->] (kdxi) -- (jsum2)
node[above,pos=0.75]{$+$};
\draw[->] (jsum2) -- (ssum)
node[right,pos=0.75]{$+$};

\draw[->] (muin) -| (musum)
node[above,pos=0.2]{$\mu_{y,ext}$}
node[right,pos=0.9]{$+$};
\draw[->] (fin) -- (fgain)
node[above,pos=0.3]{$f_{x,ext}$}
node[right,pos=1,inner sep = 0.1 em]{$\frac{1}{m}$};
\draw[->] (fgain) -| (fsum)
node[left,pos=0.9]{$+$};
	
\end{tikzpicture}}
\caption{Linearized closed-loop forward acceleration dynamics, with pitch acceleration dynamics in blue area.} \label{fig:pitchacclinear}
\end{figure*}

We consider forward and pitch movement around the hover state. The subscript $x$ indicates the forward component, \eg, $a_{x,ref} = \vect{a}_{ref}^T\vect{i}_x$, and the subscript $y$ the pitch component, \eg, ${\mu}_{y} = \vect{\mu}^T\vect{i}_y$. In hover condition, $\tau = -g$, $\theta=0$, and $\vect{\Omega} =\vect{0}_{3\times 1}$, so that \eqref{eq:vdot} and \eqref{eq:Omegadot} can be linearized to obtain
\begin{align}
a_x &= -g \theta + m^{-1}f_{x,ext},\label{eq:ax_lin}\\
J_{yy}\dot q &= \mu_y + \mu_{y,ext},\label{eq:dotqlin}
\end{align}
where $\theta$ is the pitch angle, and $J_{yy}$ is the vehicle moment of inertia about the $\vect{b}_y$-axis.
Similarly, the INDI linear acceleration control law \eqref{eq:accindi2} is linearized to obtain the error angle
\begin{equation}
-g \theta_e =-g\theta_f+ a_{x,ref} - (a_x)_f +g\theta, \label{eq:gthetac}
\end{equation}
where $-g\theta_f$ represents the forward component of the specific thrust vector, and
$(a_x)_f$ represents the filtered forward acceleration as obtained by \eqref{eq:gravcorrection}.
The commanded pitch acceleration $\alpha_c$ is obtained by taking the pitch component of \eqref{eq:PD}, as follows:
\begin{equation}\label{eq:qc}
\alpha_c = k_\theta \theta_e + k_{q} \left(q_{ref} - q_f \right) + \alpha_{ref},
\end{equation}
where $q_{ref} = -\frac{j_{x,ref}}{g}$
and
$
\alpha_{ref} = -\frac{s_{x,ref}}{g}
$ by linearization of \eqref{eq:angrateref} and \eqref{eq:angaccref}.
The scalar control gains $k_{\theta}$ and $k_{q}$ are obtained by selecting the pitch elements from the corresponding control gain matrices described in Section \ref{sec:trajtrack}.

Next, we linearize the angular acceleration and moment control laws. The four motors can be modeled collectively, as the system is linearized around the hover state where all motors have identical angular speeds. The scalar value $\omega$ refers to the deviation from the hover state motor speed $\omega_0$, or, equivalently, to half of the angular speed difference between the front and rear motor pairs.
Equating \eqref{eq:mu_c} and \eqref{eq:mutauc}, and isolating the pitch channel gives
\begin{equation}\label{eq:pitchanal_nonlin}
\omega_c = \sqrt{\left((\omega_0+\omega)^2\right)_f + J_{yy}(4l_xk_\tau)^{-1}(\alpha_c -  \alpha_f)}-\omega_0.
\end{equation}
with the factor 4 due to the number of motors.
Linearization around the hover state gives
\begin{equation}
\omega_c = \omega_f + J_{yy}k_G^{-1}(\alpha_c - \alpha_f),\label{eq:omegacLin}
\end{equation}
with the linearized control effectiveness gain
$
k_G = {8\omega_0l_xk_\tau},
$
so that
$
\mu_y = k_G\omega
$.

In order to analyze the robustness properties provided by the proposed incremental controller, it is compared to a regular, \ie, non-incremental, controller with linearized equations (cf. \eqref{eq:gthetac} and \eqref{eq:omegacLin})
\begin{equation}
\theta_{c,NI} = -\frac{a_{x,ref}}{g},
\;\;\;\;\;\;\;\;\;\;\;\;\;\;\;\;\;\;
\omega_{c,NI} = \frac{J_{yy}}{k_G}\alpha_c,
\end{equation}
where $\alpha_c$ is still given by \eqref{eq:qc} using $\theta_e = \theta_c - \theta$, and the subscript $NI$ is used to indicate the non-incremental controller.

\begin{figure*} [!h]
	\centering
	\begin{minipage}[b]{0.66\textwidth}
		\centering
		\subfloat[Position response to $f_{x,ext}$ step input.]{%
			\includegraphics[width=0.49\linewidth]{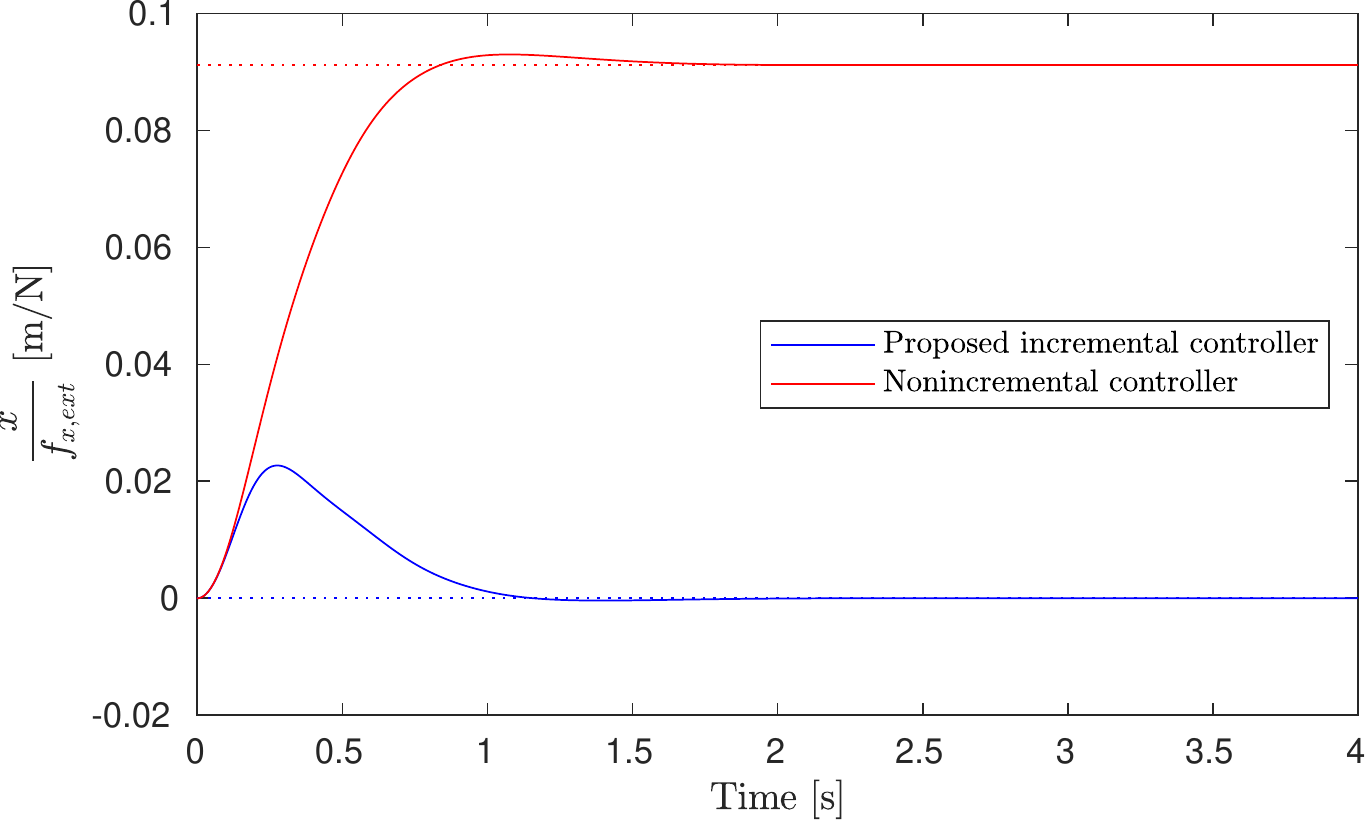}}
		\label{fig:robustness_extforce}\hfill
		\subfloat[Position response to $\mu_{y,ext}$ step input.]{%
			\includegraphics[width=0.49\linewidth]{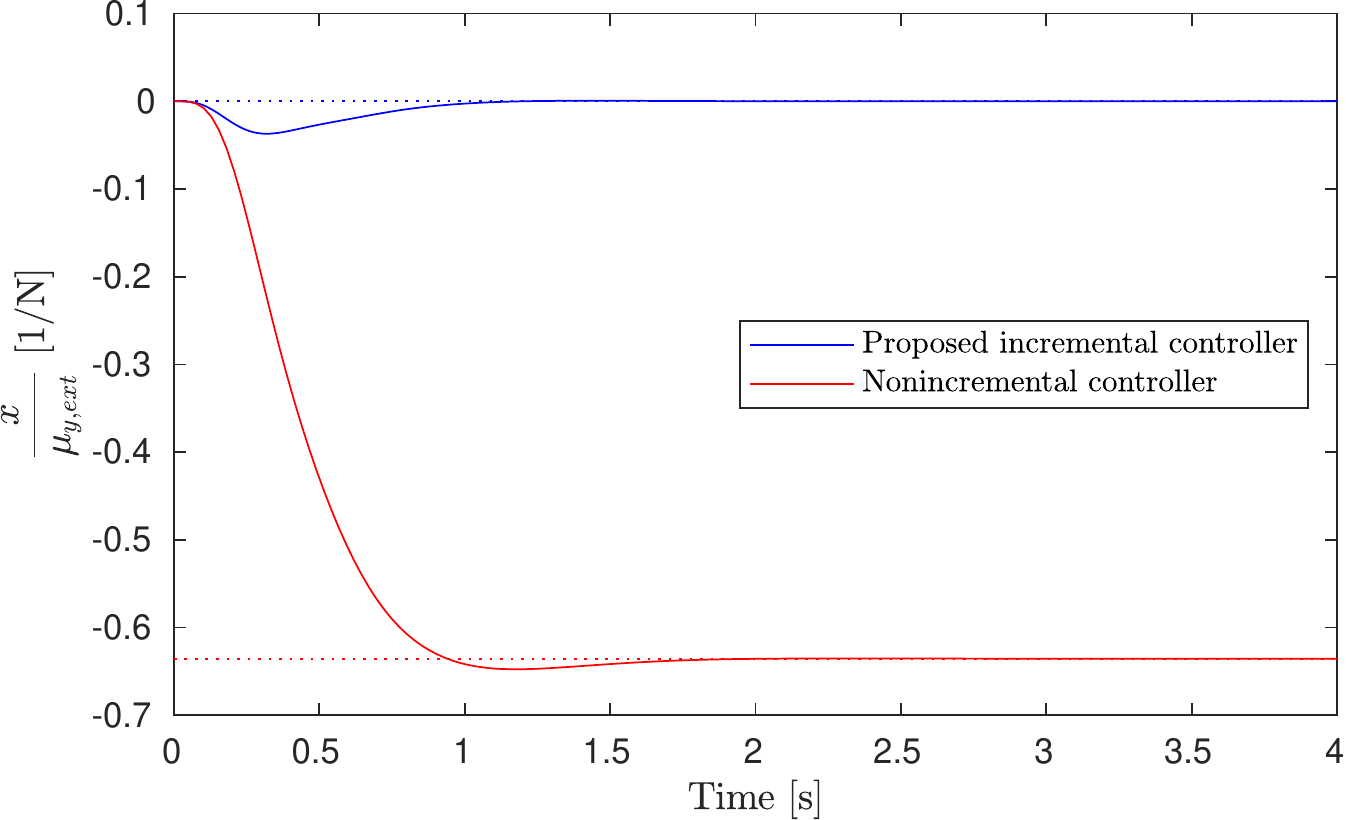}}
		\label{fig:robustness_extmoment}\hfill
		\caption{Simulated disturbance response using the proposed incremental controller, and a non-incremental controller.}
		\label{fig:robustness_extforcemoment} 
	\end{minipage}
	\hfill
	\begin{minipage}[b]{0.33\textwidth}
		\centering
		\includegraphics[width=\linewidth]{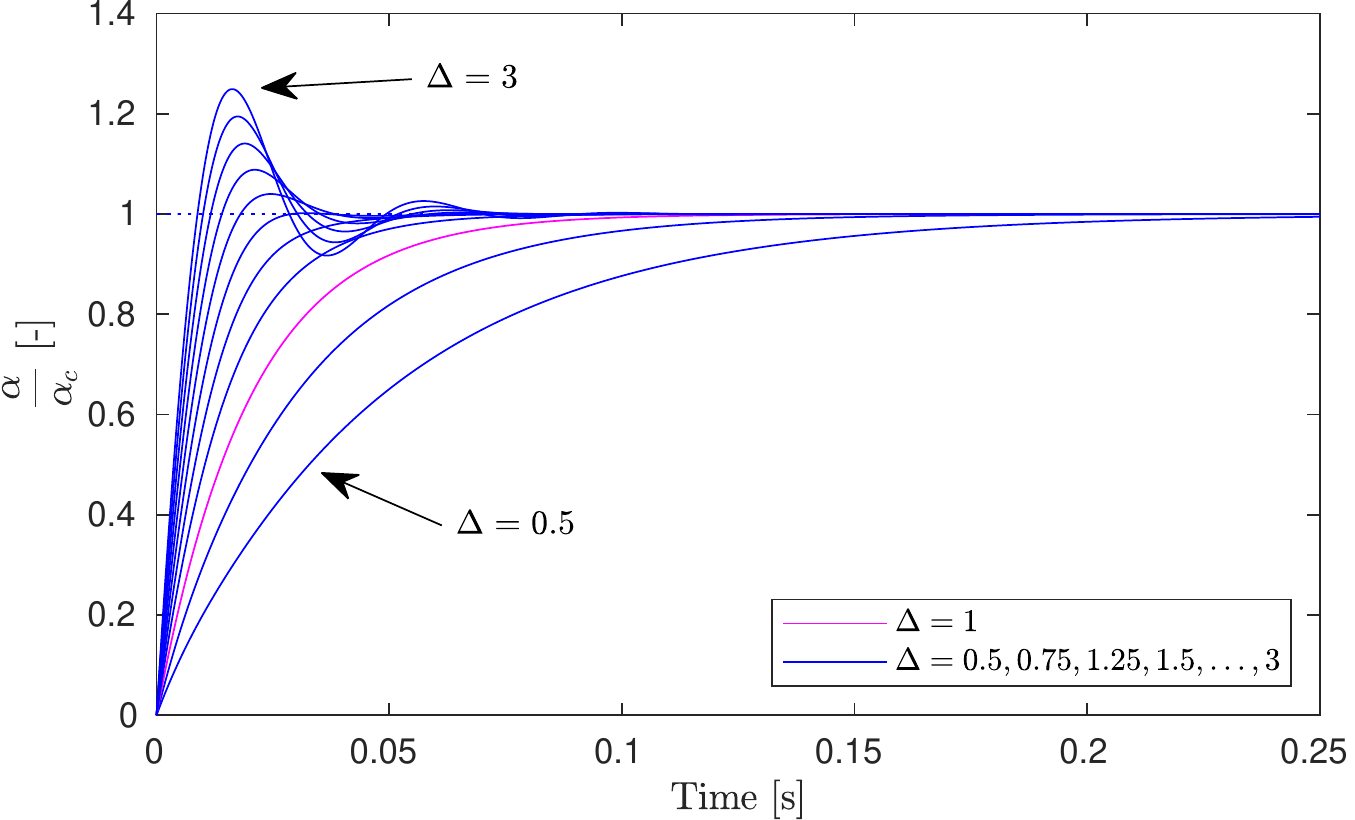}
		\caption{Simulated angular acceleration step response for various modeling errors using the proposed incremental controller.}\label{fig:angacc_modelerror}
	\end{minipage}
\end{figure*}

\subsection{Robustness against Disturbance Forces and Moments}
An overview of the resulting linearized closed-loop acceleration dynamics is given in Fig. \ref{fig:pitchacclinear}. From the blue area, we obtain the following pitch acceleration dynamics:
\begin{align}\label{eq:angacctf}
\frac{\alpha}{\alpha_c}(s)&=\frac{J_{yy}k_G^{-1}\frac{M(s)}{1-M(s)H(s)}k_GJ_{yy}^{-1}}{1+J_{yy}k_G^{-1}\frac{M(s)}{1-M(s)H(s)}k_GJ_{yy}^{-1}H(s)}=M(s),\\
\frac{\alpha}{\mu_{y,ext}}(s) &= \frac{J_{yy}^{-1}}{1+\frac{H(s)M(s)}{1-H(s)M(s)}}=J_{yy}^{-1}\left(1-H(s)M(s)\right)\label{eq:angacctf_muext}
\end{align}
with $\alpha(s)$ the pitch acceleration, \ie, $\alpha(s)=sq(s)=s^2\theta(s)$.
The LPF transfer function is denoted by $H(s)$, \eg, $\frac{\alpha_f}{\alpha}(s) = H(s)$, and the motor (control) dynamics are denoted by $M(s)$, \ie, $\frac{\omega}{\omega_c}(s) = M(s)$.
In \eqref{eq:angacctf}, we observe that the closed-loop angular acceleration dynamics are solely determined by the motor dynamics~\cite{smeur2015adaptive}. Hence, the aggressiveness of trajectories that can be tracked is theoretically limited by only the bandwidth of the motor response.
This is also the case for a non-incremental version of the controller.

The disturbance moment $\mu_{y,ext}$ is fully counteracted using incremental control based on the two feedback loops in the blue shaded area of Fig. \ref{fig:pitchacclinear}:
the expected angular acceleration from the motor speeds, \ie, $\frac{k_G}{J_{yy}}\omega_f$,
and the measured angular acceleration $\alpha_f$, which includes the effects of the disturbance moment.
As shown in \eqref{eq:angacctf_muext}, the counteraction depends on $H(s)$ and $M(s)$ so that the ability to reject disturbances is limited by the bandwidth of both the LPFs and the motors.
To the contrary, in a non-incremental controller the $\alpha_f$ and $\omega_f$ feedback loops are not present, so that $\alpha= J^{-1}_{yy}\mu_{y,est}$ (cf. \eqref{eq:angacctf_muext}).
The disturbance moment now propagates undamped to the attitude and position control loops, as there is no closed-loop angular acceleration control that directly evaluates the moments acting on the vehicle.

We obtain similar results for the disturbance force $f_{x,ext}$, which is corrected for incrementally using the difference between the acceleration due to thrust, \ie, $-g\theta_f$, and the true acceleration including the disturbance force, \ie, $(a_x)_f$.
All in all, the proposed incremental controller maintains identical nominal reference tracking performance for both angular and linear accelerations, while achieving superior disturbance rejection of external moments and forces when compared to the non-incremental controller.

In order to evaluate the effect of the disturbance force and moment on the position tracking error, we close the loop around Fig. \ref{fig:pitchacclinear} using the position and velocity controller given by \eqref{eq:pd}.
Figure \ref{fig:robustness_extforcemoment} shows the resulting step responses for both incremental and non-incremental control. The response was simulated using the platform-independent control gains given in Table \ref{tab:gains}, a second-order Butterworth filter with cut-off frequency equal to 188.5 \si{rad/s} (30 \si{Hz}), and the first-order motor model given by \eqref{eq:motordyn}
with $\uptau_m$ set to 20 \si{ms}. It can be seen that the proposed incremental controller is able to counteract the disturbances and reaches zero steady-state error, while the non-incremental controller is unable to do so.
In order to null the steady-state errors due to force and moment disturbances, integral action must be added to the non-incremental controller. This is not necessary in the case of INDI, so that our proposed control design is able to quickly and wholly counteract disturbance forces and moments, while avoiding the negative effects that integral action typically has on the tracking performance, \eg, degraded stability, and increased overshoot and settling time.

\begin{table}
	\centering
	\caption{Trajectory tracking controller gains.}
	\label{tab:gains}
	\begin{tabular}{rl}
		\hline
		Gain & Value\\
		\hline
		$\vect{K_x}$ & $\operatorname{diag}\left([18 \; 18 \; 13.5]\right)$\\
		$\vect{K_v}$ & $\operatorname{diag}\left([7.8 \; 7.8 \; 5.9]\right)$\\
		$\vect{K_a}$ & $\operatorname{diag}\left([0.5 \; 0.5 \; 0.3]\right)$\\
		$\vect{K_\xi}$ & $\operatorname{diag}\left([175 \; 175 \; 82]\right)$\\
		$\vect{K_{\dot \xi}}$ & $\operatorname{diag}\left([19.5 \; 19.5 \; 19.2]\right)$\\
		\hline
	\end{tabular}
\end{table}	

\begin{figure*} [th]
	\centering
	\begin{minipage}[b]{0.66\textwidth}
		\centering
		\subfloat[Using the proposed incremental controller.]{%
			\includegraphics[width=0.49\linewidth]{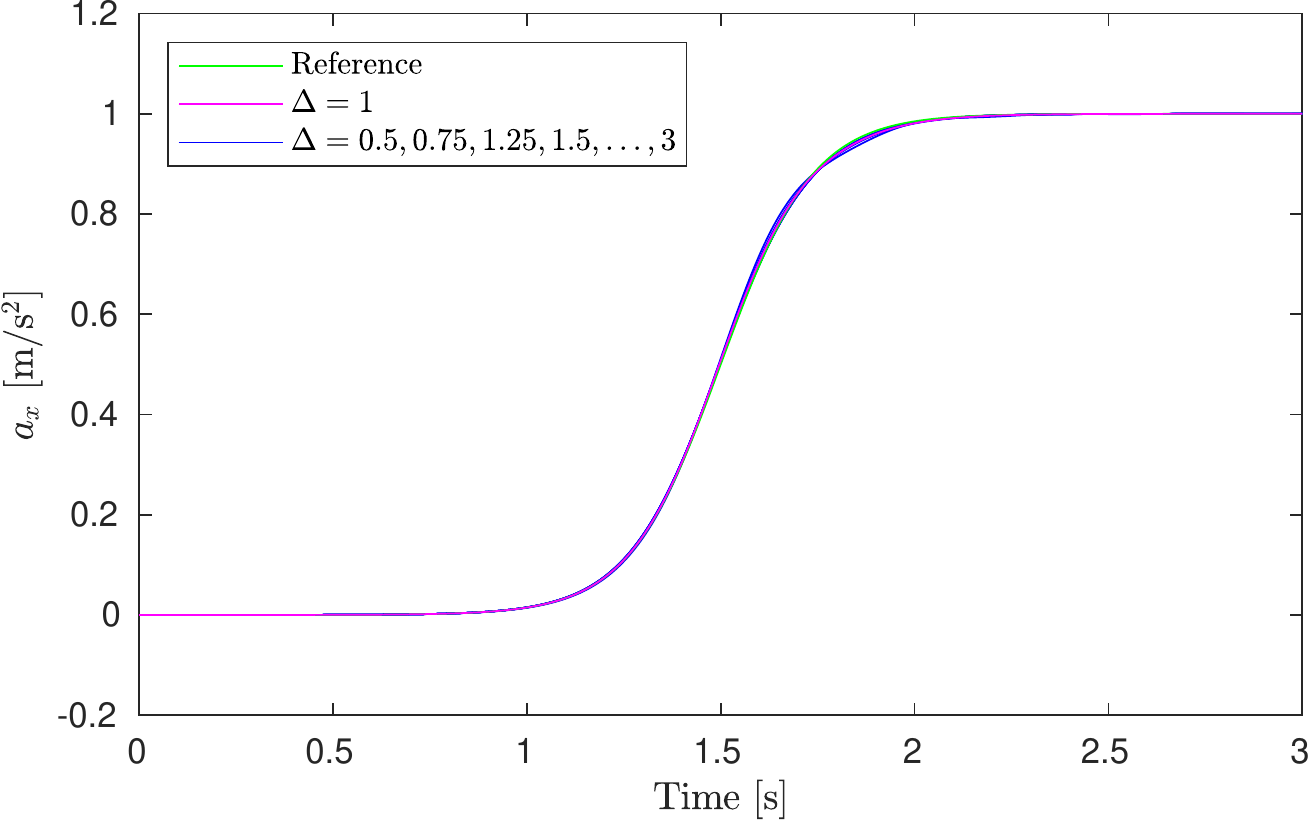}}
		\label{fig:linacc_modelingerror_proposed}\hfill
		\subfloat[Using a non-incremental controller.]{%
			\includegraphics[width=0.49\linewidth]{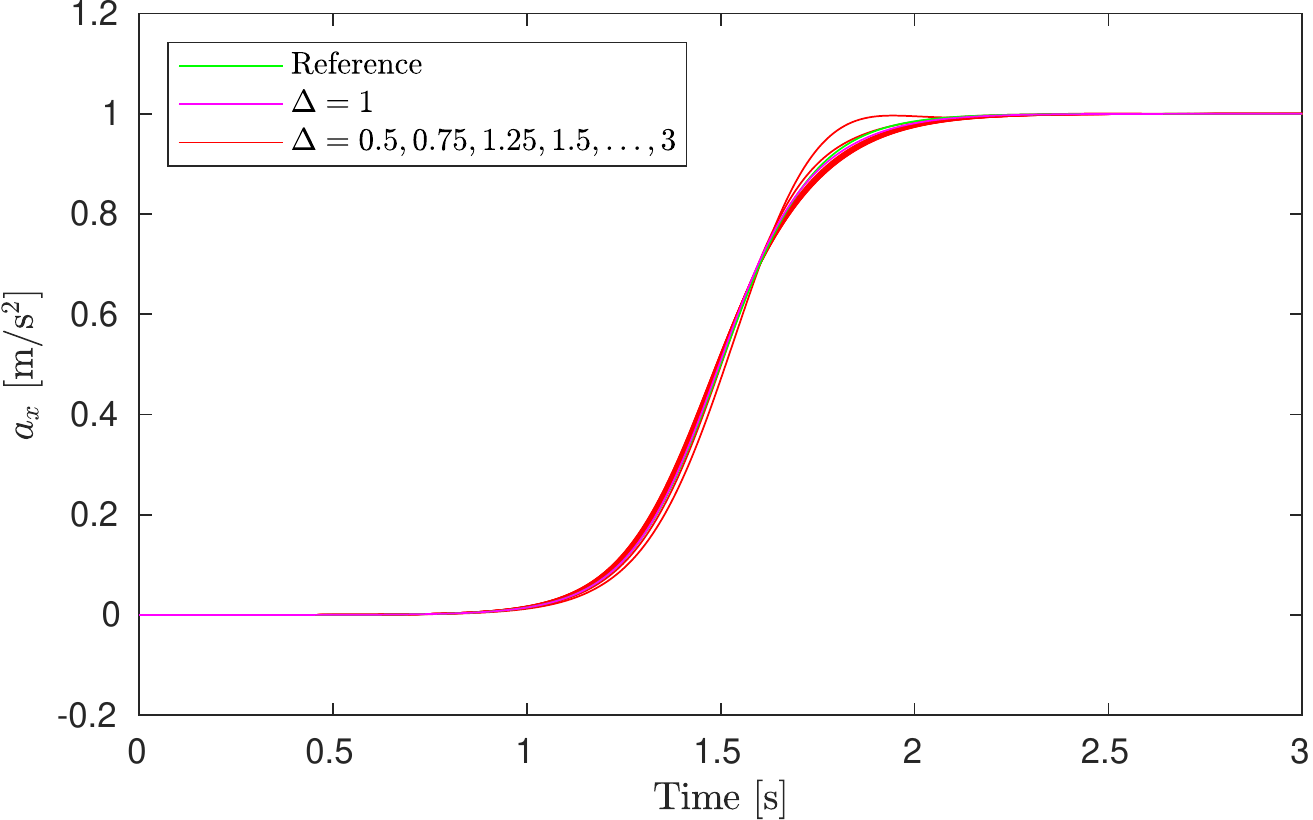}}
		\label{fig:linacc_modelingerror_nonincr}\hfill
		\caption{Simulated linear acceleration tracking response for various modeling errors.\vspace{2.2em}}
		\label{fig:linacc_modelingerror} 
	\end{minipage}
	\hfill
	\begin{minipage}[b]{0.32\textwidth}
		\centering
		\includegraphics[width=\linewidth]{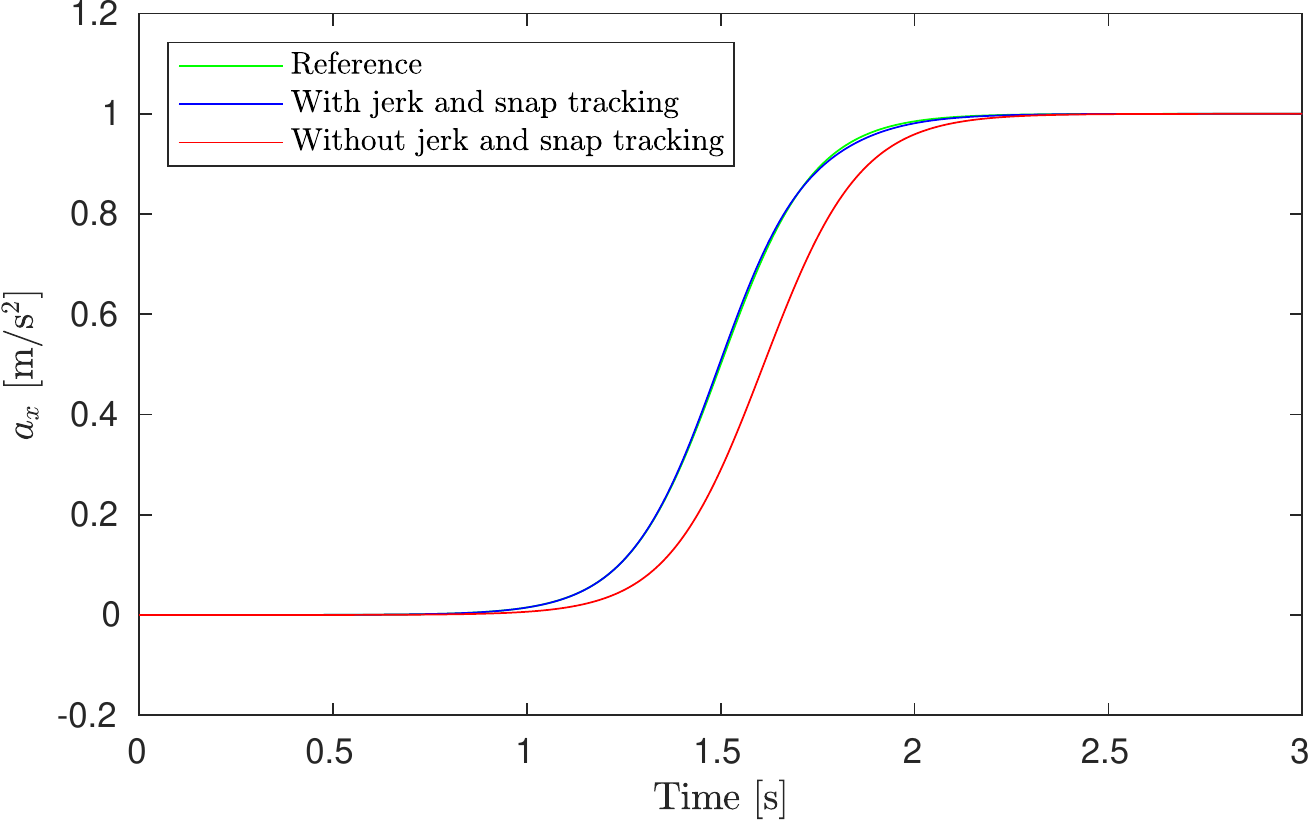}
		\caption{Simulated linear acceleration tracking response using the proposed controller with and without jerk and snap tracking.}\label{fig:acc_reponse}
	\end{minipage}
\end{figure*}

\begin{table*}[!b]
	\begin{minipage}[t]{0.49\textwidth}
		\centering
		\caption{3D trajectory tracking performance for experiments with forward yaw and constant yaw.}
		\vspace{1.5em}
		\label{tab:trackingperformance-comptraj}
		\begin{tabular}{lcc}
			\hline
			& Forward yaw & Constant yaw\\
			\hline
			RMS $\|\vect{x}-\vect{x}_{ref}\|_2$ [cm] &6.6 & 6.1\\
			max $\|\vect{x}-\vect{x}_{ref}\|_2$ [cm]&10.8 &11.9 \\
			RMS $|\psi-\psi_{ref}|$ [deg]& 5.1 & 1.9\\
			max $|\psi-\psi_{ref}|$ [deg]& 12.8& 6.4\\
			RMS $\|\vect{v}\|_2$ [m/s]&6.8 & 5.6 \\
			max $\|\vect{v}\|_2$ [m/s]&12.9& 11.3\\
			RMS $\|\vect{a} - g\vect{i}_z\|_2$ [m/s\textsuperscript{2}] &14.4& 12.5 \\
			max $\|\vect{a} - g\vect{i}_z\|_2$ [m/s\textsuperscript{2}]& 20.8& 20.0\\
			\hline
		\end{tabular}
	\end{minipage} \hfill
	\begin{minipage}[t]{0.49\textwidth}
		\centering
		\caption{Roulette curve trajectory tracking performance for: \textrm{\normalfont \textbf{(i)}} the proposed controller; \textrm{\normalfont \textbf{(ii)}} jerk and snap tracking disabled; and \textrm{\normalfont \textbf{(iii)}} drag plate attached.}
		\label{tab:trackingperformance}
		\begin{tabular}{lccc}
			\hline
			& (i) & (ii) & (iii) \\
			\hline
			RMS $\|\vect{x}-\vect{x}_{ref}\|_2$ [cm] & 9.0& 16.8& 7.6\\
			max $\|\vect{x}-\vect{x}_{ref}\|_2$ [cm] & 14.3 &28.9 & 14.2\\
			RMS $\psi$ [deg] & 1.8& 5.3& 12.6\\
			max $|\psi|$ [deg] & 5.3& 14.9& 51.7\\
			RMS $\|\vect{v}\|_2$ [m/s] & 3.7& 4.3& 3.8\\
			max $\|\vect{v}\|_2$ [m/s] & 7.3& 8.2& 7.7\\
			RMS $\|\vect{a} - g\vect{i}_z\|_2$ [m/s\textsuperscript{2}]& 14.0 & 15.2& 14.2 \\
			max $\|\vect{a} - g\vect{i}_z\|_2$ [m/s\textsuperscript{2}] &19.1 & 21.3& 20.4\\
			\hline
		\end{tabular}
	\end{minipage} 
\end{table*}

\subsection{Robustness against Modeling Errors}
The proposed control design requires only few vehicle-specific parameters.
Nonetheless, it is desirable that tracking performance is maintained if inaccurate parameters are used, \eg, because control effectiveness data obtained from static tests may not be representative for the entire flight envelope.
The linearized control equations described above incorporate the ratio of the moment of inertia $J_{yy}$ and the linearized control effectiveness $k_G$. We denote the values used in the controller $\bar J_{yy}$ and $\bar k_G$, and define the modeling error $\Delta$ such that
\begin{equation}
\frac{\bar J_{yy}}{\bar k_G}= \Delta \frac{ J_{yy}}{ k_G}.
\end{equation}
This leads to the following pitch acceleration dynamics for the proposed incremental NDI controller, and the non-incremental controller described above:
\begin{equation}
\frac{\alpha}{\alpha_c}(s) = \frac{\Delta M(s)}{(\Delta - 1)H(s)M(s) + 1},\label{eq:Deltaacc}
\end{equation}
\begin{equation}
\frac{\alpha}{\alpha_c}_{NI}(s) =\Delta M(s).\label{eq:Deltaacc_1}
\end{equation}
It can be seen that the error acts as a simple gain in the non-incremental controller, leading to an incorrect angular acceleration. On the contrary, the proposed incremental controller compares the expected angular acceleration from the motor speeds, \ie, $\frac{\bar k_G}{\bar J_{yy}}\omega_f$, to the measured angular acceleration, \ie, $\alpha_f$, to implicitly correct for the modeling error. The corresponding angular acceleration responses for several values of $\Delta$ are shown in Fig. \ref{fig:angacc_modelerror}. The figure shows that the modeling error affects the transient response, but that the incremental controller is able to correct for it and quickly reaches the commanded acceleration value even for very large model discrepancies.

In order to assess the effect of modeling errors on acceleration tracking, we simulate the time response to the following acceleration reference:
\begin{equation}\label{eq:accref}
a_{x,ref}(t) = \frac{1}{2}\tanh\left(\frac{4}{3}\pi t - 2 \pi\right) + \frac{1}{2},
\end{equation}
which is $C^2$, \ie, the corresponding jerk and snap signals are continuous, and has boundary conditions $a_{x,ref}(0)=j_{x,ref}(0)=s_{x,ref}(0)=j_{x,ref}(3)= s_{x,ref}(3) = 0$ and $a_{x,ref}(3)=1$ \si{m/s\textsuperscript{2}}. The responses for various values of $\Delta$ are shown in Fig. \ref{fig:linacc_modelingerror}. It can be seen that the incremental controller is able to accurately track the reference signal even when large modeling errors are present. When non-incremental control is used, the tracking performance declines more severely with growing modeling error.

\subsection{Jerk and Snap Tracking}
Jerk and snap tracking is a crucial aspect of the proposed controller design that enables tracking of fast-changing acceleration references.
It is embodied by the feedforward terms $k_q s$ and $s^2$ in the nominator of the acceleration response transfer function
\begin{equation}
\frac{a_x}{a_{x,ref}}(s) = \frac{M(s)\left(s^2+k_q s+k_\theta\right)}{s^2+k_q H(s)M(s)s+k_\theta M(s)}.
\end{equation}
These feedforward terms add two zeros to the closed-loop transfer function.
These zeros --- in combination with the LPF --- act essentially as a lead compensator and help improve the transient response of the system.
Effective placement of the zeros through tuning of $k_{q}$ leads to improved tracking of a rapidly changing acceleration input signal, \eg, during aggressive flight maneuvers.

Figure \ref{fig:acc_reponse} shows the simulated acceleration responses with and without jerk and snap tracking to the reference signal defined in \eqref{eq:accref}. It can be seen that the inclusion of jerk and snap tracking causes a faster response, resulting in more accurate acceleration tracking. In the next section, we show that the improvement is also achieved in practice.

\begin{figure*}
	\centering
	
	\subfloat[Forward yaw.]{\includegraphics[width=0.42\linewidth,valign=c]{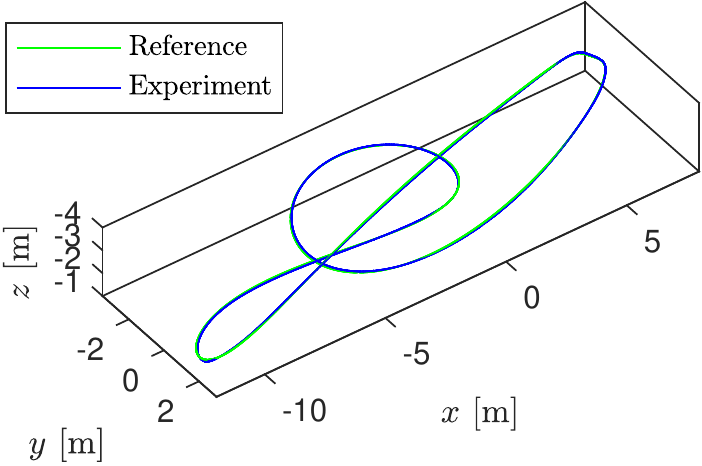}\label{fig:3d-traj1}}\hspace{7em}
	\subfloat[Constant yaw.]{%
		\includegraphics[width=0.42\linewidth,valign=c]{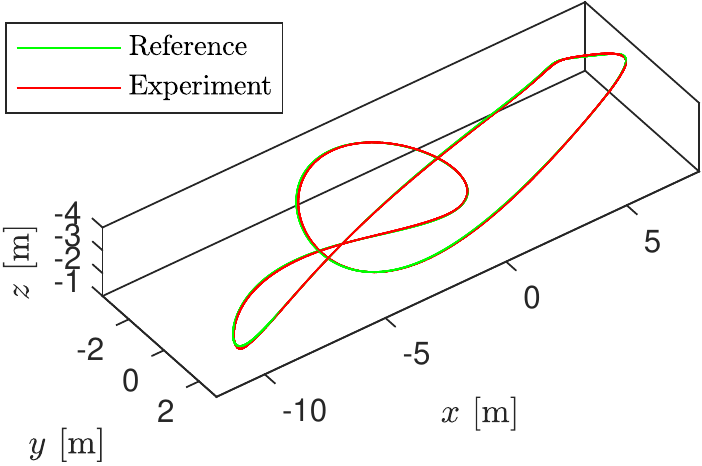}\label{fig:3d-traj2}%
	}% \quad
	
	\caption{Experimental flight results for 3D trajectory.}\label{fig:3d-traj}
\end{figure*}
\begin{figure*}
	\centering
	
	\subfloat[Euclidean norm of position error.]{\includegraphics[width=0.49\linewidth,valign=c]{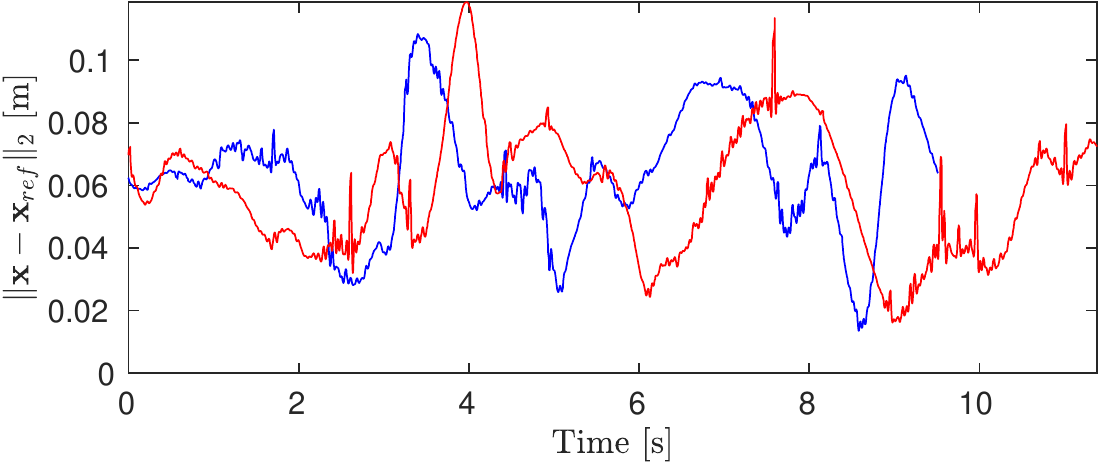}\label{fig:rollercoaster-poserror}}\hfill
	\subfloat[Yaw error.]{%
		\includegraphics[width=0.49\linewidth,valign=c]{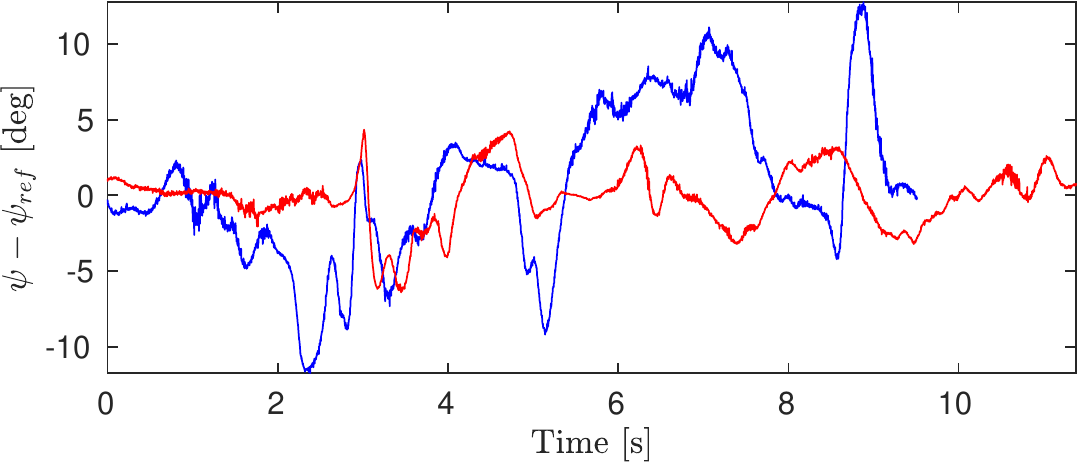}\label{fig:rollercoaster-psierror}%
	}% \quad
	
	\subfloat[Euclidean norm of velocity.]{\includegraphics[width=0.49\linewidth,valign=c]{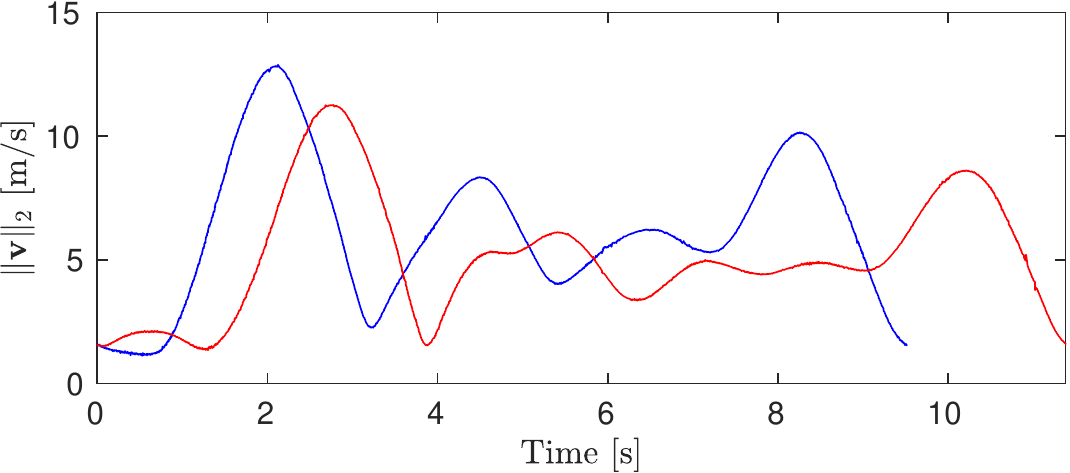}\label{fig:rollercoaster-speed}}\hfill
	\subfloat[Euclidean norm of acceleration.]{%
		\includegraphics[width=0.49\linewidth,valign=c]{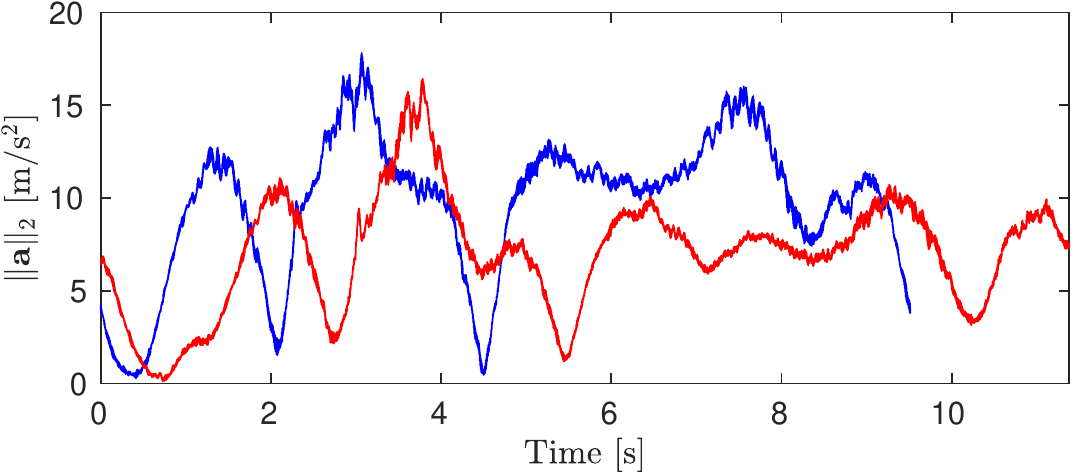}\label{fig:rollercoaster-accel}%
	}% \quad
	
	\caption{Experimental flight results for 3D trajectory: forward yaw (blue), and constant yaw (red).}\label{fig:rollercoaster}
\end{figure*}

\section{Experimental Results}\label{sec:experiments}
In this section, experimental results for high-speed, high-acceleration flight are presented.
A video of the experiments is available at \url{https://youtu.be/K15lNBAKDCs}.
We evaluate the performance of the trajectory tracking controller on two trajectories that include yawing, tight turns with acceleration up to over 2\si{g}, and high-speed straights at up to 12.9 \si{m/s}.
Furthermore, we examine the effect of the feedforward inputs based on the reference trajectory jerk and snap.
We establish the independence of any model-based drag estimate by attaching a drag-inducing cardboard plate that more than triples the frontal area of the vehicle. Robustness against external disturbance forces is further displayed by pulling on a string attached to the quadcopter in hover.
Finally, we compare the proposed nonlinear INDI angular acceleration control to its linearized counterpart.

\begin{figure*}[!t]
	\centering
	
	\subfloat[Position.]{\includegraphics[width=0.49\linewidth,valign=c]{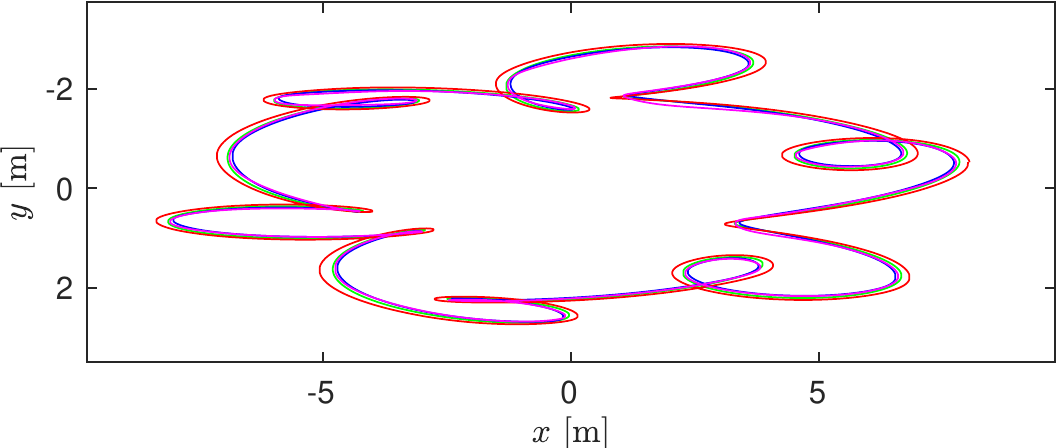}}\label{fig:traj}\hfill
	\subfloat[Euclidean norm of position error.]{\includegraphics[width=0.49\linewidth,valign=c]{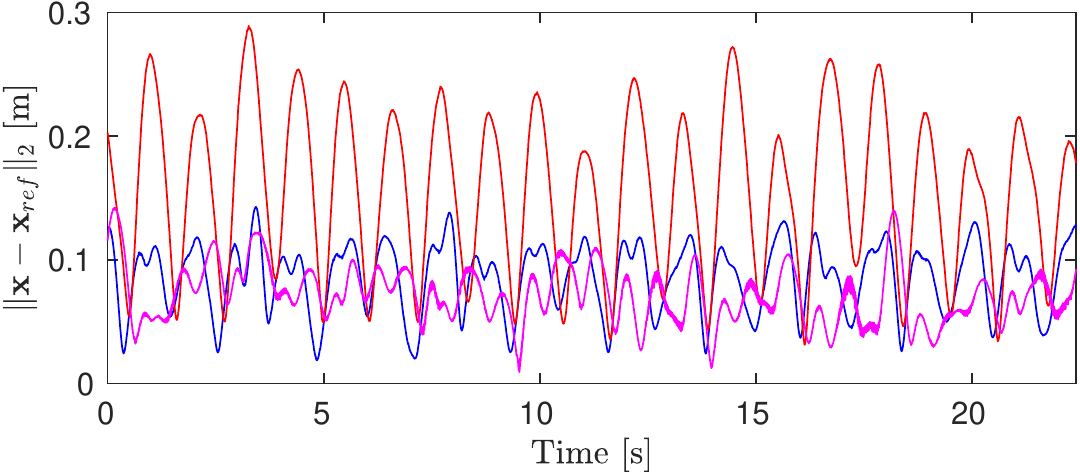}}\label{fig:fig8_poserror}
	
	\subfloat[Euclidean norm of velocity.]{\includegraphics[width=0.49\linewidth,valign=c]{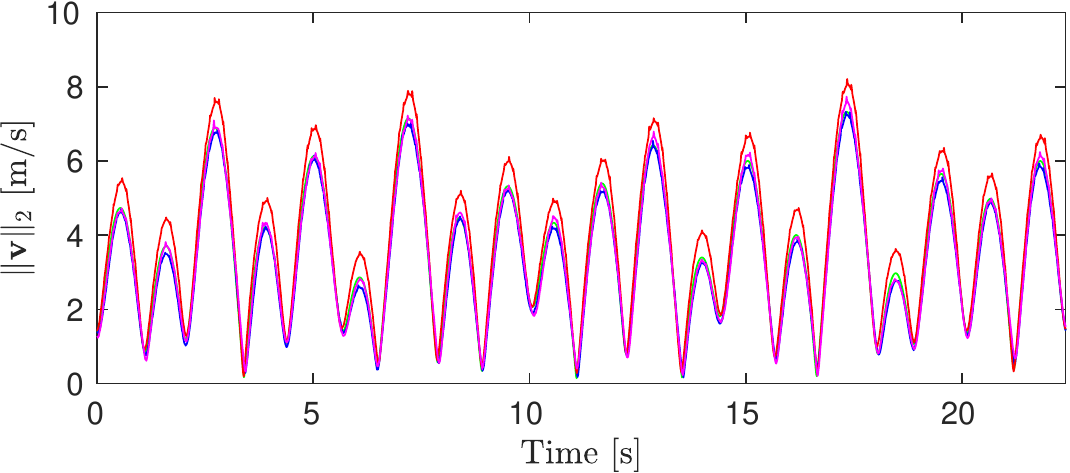}}\label{fig:vel}\hfill
	\subfloat[Euclidean norm of acceleration.]{\includegraphics[width=0.49\linewidth,valign=c]{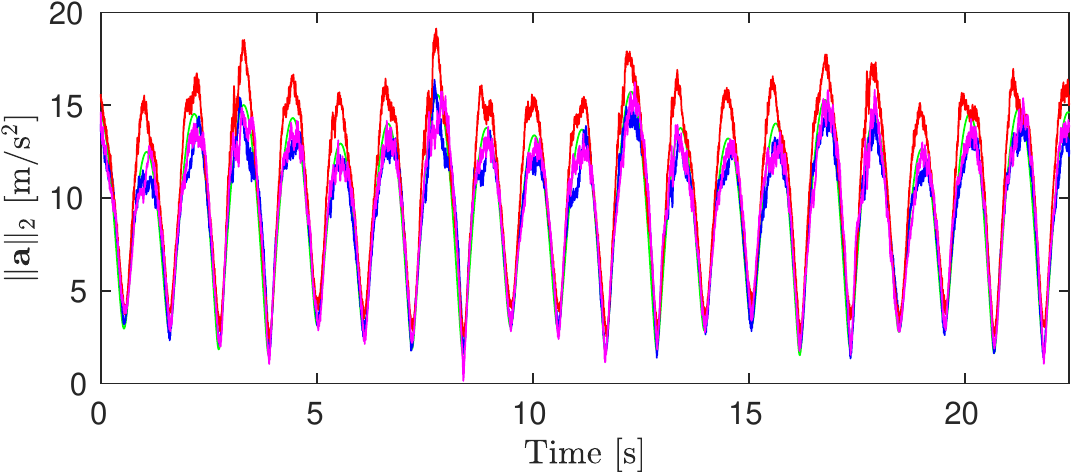}}\label{fig:acc}
	
	\caption{Experimental flight results for roulette curve trajectory: reference trajectory (green), proposed controller (blue), without jerk and snap tracking (red), and with drag plate attached (magenta).}\label{fig:velacc}
\end{figure*}

\subsection{Experimental Setup}
Experiments were performed in an indoor flight room using the quadcopter shown in Fig. \ref{fig:refsystems}.
The quadrotor body is machined out of carbon fiber composite with balsa wood core.
The propulsion system consists of T-Motor F35A ESCs and F40 Pro II Kv 2400KV motors with Gemfan Hulkie 5055 propellers.
Adjacent motors are mounted 18 \si{cm} apart.
The quadcopter is powered by a single 4S LiPo battery.
Its total flying mass is 609 \si{g}.

Control computations are performed at 2000 \si{Hz} using an onboard STM32H7 400 \si{MHz} microcontroller running custom firmware.
On this platform, the total computation time of a control update at 32-bit floating point precision is 16 $\mu$\si{s}.
Linear acceleration and angular rate measurements are obtained from an onboard Analog Devices ADIS16477-3 IMU at 2000 \si{Hz}, while position, velocity, and orientation measurements are obtained from an OptiTrack motion capture system at 360 \si{Hz} with an average latency of 18 \si{ms}. The latency is corrected for by propagating motion capture data using integrated IMU measurements.
Motor speed measurements are obtained from the optical encoders at approximately 5000 \si{Hz}.
The motor speed and IMU measurements are low-pass filtered using a software second-order Butterworth filter with cutoff frequency 188.5 \si{rad/s} (30 \si{Hz}).

The platform-independent controller gains listed in Table \ref{tab:gains} were used.
Additionally, the controller requires several platform-specific parameters, namely:
vehicle mass $m$, moment of inertia $\vect{J}$, motor time constant $\uptau_m$, control effectiveness matrices $\vect{G}_1$ and $\vect{G}_2$, and the gain and polynomial fit used by the motor speed controller.
We obtained control effectiveness data from static tests.
In experiments, it was found that using the controller on a different quadcopter
(with different dynamic properties, inertial sensors, and propulsion system) required no changes to controller algorithms or gains.
After updating only the aforementioned platform-specific parameters, the controller performed without loss of tracking accuracy.

\subsection{Evaluation of Proposed Controller}
In this section, we evaluate the performance of the trajectory tracking controller on two trajectories: a 3D trajectory that includes a high-speed straight and fast turns, and a roulette curve trajectory consisting of fast successive turns resulting in high jerk and snap.
The 3D trajectory is generated from a set of waypoints using the method described in \cite{ry_rss2020}.
The trajectory is flown with two yaw references: forward yaw, \ie, with the $\vect{b}_x$-axis in the velocity direction, and constant yaw set to zero.
Figure \ref{fig:3d-traj} shows the corresponding reference trajectories, along with experimental results.
The forward yaw trajectory is flown in slightly shorter time.
Performance data for both trajectories are given in Table \ref{tab:trackingperformance-comptraj} and shown in Fig. \ref{fig:rollercoaster}.
Over the forward yaw trajectory a maximum speed of 12.9 \si{m/s} is achieved, while the RMS tracking error is limited to 6.6 \si{cm}.
The vehicle attains a maximum proper acceleration of 20.8 \si{m/s}\textsuperscript{2} (2.12\si{g}).
Similar values can be observed for the constant yaw trajectory.
The most significant difference is a reduction in yaw tracking error from an RMS value of 5.1 \si{deg} to 1.9 \si{deg} and from a maximum value of 13 \si{deg} to 6.4 \si{deg}.

The second, roulette curve trajectory is defined as
\begin{equation}
\vect{\sigma}_{ref}(t) = \left[\begin{array}{c}
r_1 \cos k_1 t + r_2 \cos k_2 t + r_3 \sin k_3 t\\
r_4 \sin k_1 t + r_3 \sin k_2 t + r_5 \cos k_3 t\\
r_{z}\\
0
\end{array}\right]\label{eq:lemntraj}
\end{equation}
with $r_{1} = $ 6 \si{m}, $r_{2} = $ 1.8 \si{m}, $r_{3} = $ 0.6 \si{m}, $r_{4} = $ -2.25 \si{m}, $r_{5} = $ -0.3 \si{m}, $r_{6} = $ -0.45 \si{m}, $k_{1} = $ 0.28 \si{rad/s}, $k_{2} = $ 2.8 \si{rad/s}, $k_{3} = $ 1.4 \si{rad/s}, and $r_{z}$ a constant offset.
The trajectory, shown in Fig. \ref{fig:velacc}(a), contains fast, successive turns.
Accurate tracking is particularly demanding as it requires fast changes in acceleration, \ie, large jerk and snap, requiring high angular rates and angular accelerations. A single lap is traversed in 22.4 \si{s}.
The position tracking error is shown in blue in Fig. \ref{fig:velacc}(b), and tracking performance metrics are given in the first column of Table \ref{tab:trackingperformance}.
Comparison of the position tracking error to the values in Table \ref{tab:trackingperformance-comptraj} confirms that the controller achieves consistent performance across trajectories.
Due to its arduous nature, the roulette curve trajectory is particularly suitable to expose differences in tracking performance.
Therefore, we use the trajectory defined by \eqref{eq:lemntraj} to examine several modifications in subsequent sections. In all cases, the trajectory parameters are identical to those given above.

\begin{figure}[t]
	\centering
		\includegraphics[width=0.98\linewidth]{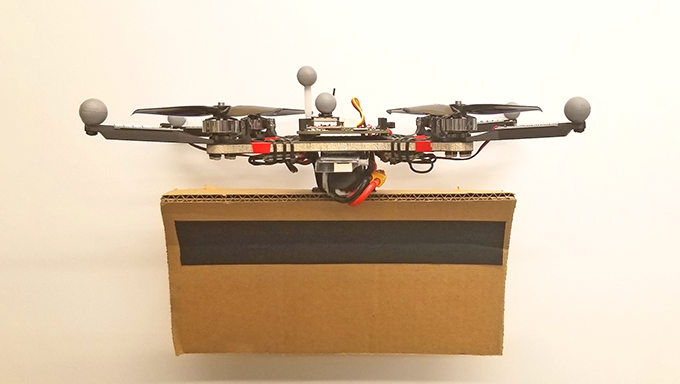}
		\caption{Quadrotor with 16 \si{cm} $\times$ \si{32} cm cardboard drag plate.}\label{fig:dragspoiler}
\end{figure}

\begin{figure}
	\centering
		\includegraphics[width=\linewidth]{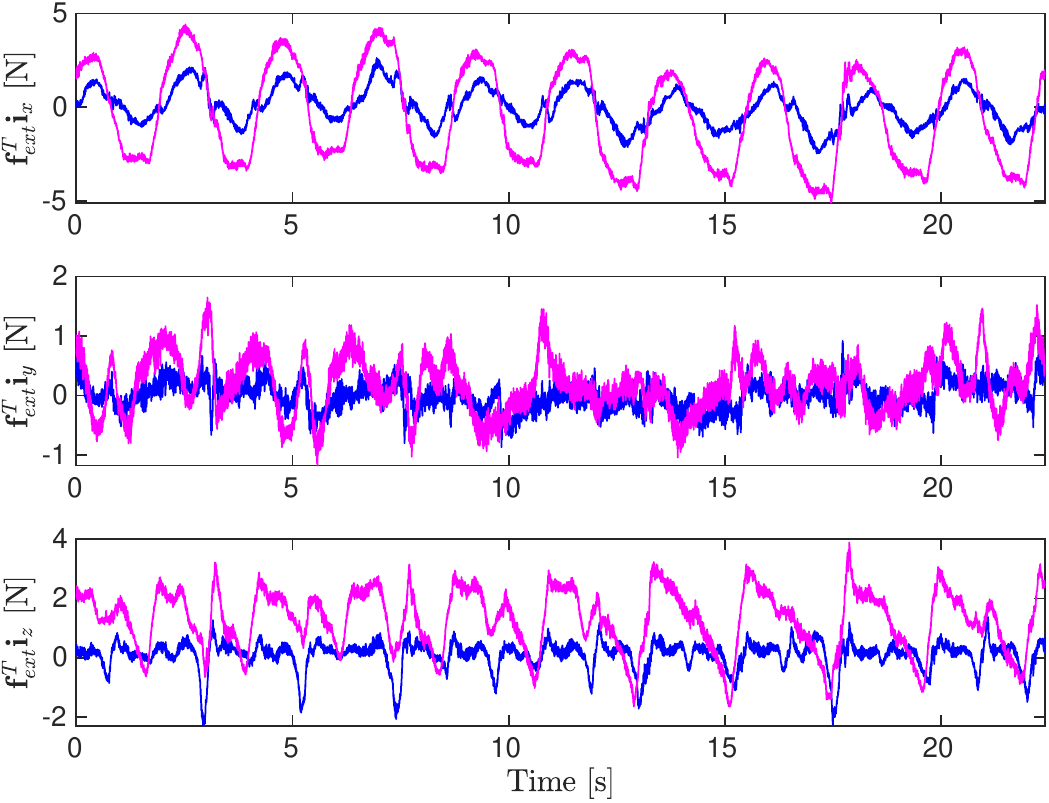}
		\caption{Estimated external disturbance force for roulette curve trajectory: proposed controller (blue), and with drag plate attached (magenta).}\label{fig:fig8-fext}
\end{figure}

\subsection{Jerk and Snap Tracking}
The red curves in Fig. \ref{fig:velacc} correspond to our proposed control design, but with jerk and snap tracking disabled, \ie, $\vect{\Omega}_{ref}=\vect{\dot \Omega}_{ref} = \vect{0}_{3\times1}$. 
Examination of the figures shows the significant improvement in trajectory tracking performance obtained through the tracking of the jerk and snap feedforward terms.
This observation is confirmed by comparing the first two columns of Table \ref{tab:trackingperformance}.
It can be seen that the RMS position tracking error increases from 9.0 \si{cm} to 16.8 \si{cm} when jerk and snap tracking are disabled.
In Section \ref{sec:analysis}, it was shown that lead compensation provided by jerk and snap tracking results in improved performance when tracking fast-changing acceleration commands. This effect can also be observed in Fig. \ref{fig:velacc}. It can be seen that the system response has less overshoot when jerk and snap tracking are enabled, conform the analytical response of the linearized system.

\subsection{Increased Aerodynamic Drag}
The magenta curves in Fig. \ref{fig:velacc} correspond to the trajectory tracking controller as described in this paper, but using the quadcopter with attached drag plate. The drag plate is a 16 \si{cm} $\times$ 32 \si{cm} cardboard plate that is attached to the bottom of the quadrotor, as shown in Fig. \ref{fig:dragspoiler}. The plate more than triples the frontal surface area of the quadrotor, and as such has a significant effect on the aerodynamic force and moment that act on the vehicle, especially during high-speed flight, and fast pitch and yaw motion. The flight controller is not adapted in any way to account for either these aerodynamic effects, or the changes in mass and moment of inertia.

Comparison of columns (i) and (iii) in Table \ref{tab:trackingperformance} shows that the drag plate does not significantly affect position tracking performance.
Yaw tracking performance is also consistent, except when the drag plates generates an external yaw moment that causes motor speed saturation and very large momentary yaw tracking error.
The consistent tracking performance demonstrates the robustness property of INDI. Controllers that depend on the estimation of drag forces based on velocity, such as \cite{faessler2018differential} and \cite{svacha2017improving}, may suffer from much larger loss of tracking performance when the aerodynamic properties of the vehicle are modified. Instead of depending on a model-based drag estimate, INDI counteracts the disturbance force and moment by sensor-based incremental control.
The controller implicitly estimates the external force by \eqref{eq:fext}. In Fig. \ref{fig:fig8-fext}, it can be seen that the drag plate has a significant effect on the external disturbance force: its estimated magnitude is approximately tripled. In order to counteract the greater external force, commanded thrust and vehicle pitch increase when the drag plate is attached.

\subsection{Nonlinear Control Effectiveness Inversion}\label{sec:nonlinindiexp}
We also compare our proposed nonlinear inversion of the control effectiveness \eqref{eq:mutauc}, with linearized INDI as presented in \cite{smeur2017cascaded}. In the latter case, control moment and thrust commands are tracked using linearized inversion of \eqref{eq:mutau}, as follows:
\begin{multline}\label{eq:linearizedINDI}
\vect{\omega}_c=\vect{\omega}_f+\left(2\vect{G}_1+\Delta t^{-1}\vect{G_2}\right)\\
\left(\left[\begin{array}{c}
\vect{\mu}_c-\vect{\mu}_f\\
T_c-T_f
\end{array}\right]+\Delta t^{-1}\vect{G}_2B(\vect{\omega}_c-\vect{\omega}_f)\right),
\end{multline}
where $B$ is the one-sample backshift operator and $\Delta t$ is the controller update interval.
This linearized inversion does not take into account local nonlinearity of \eqref{eq:mutau}, nor does it consider the transient response of the motors. Therefore, nonlinear inversion of \eqref{eq:mutauc} --- as described in Section \ref{sec:motorspeedctrl} --- theoretically results in improved tracking of the angular acceleration command and thereby in improved trajectory tracking performance.

In experimental flights we found that the difference between nonlinear and linearized inversion does not lead to significant differences in tracking performance for our quadrotor system.
However, we found that the failure to properly consider the transient response of the motors in \eqref{eq:linearizedINDI} can be detrimental for controller performance.
In particular, if the motor time constant $\uptau_m$ and the controller interval $\Delta t$ differ greatly, this may result in fast yaw oscillations.
Consideration of the motor time constant $\uptau_m$, as in \eqref{eq:mutauc}, resolves this issue.

\subsection{Hover with Disturbance Force}
\begin{figure}[b]
	\centering
	\includegraphics[width=\linewidth]{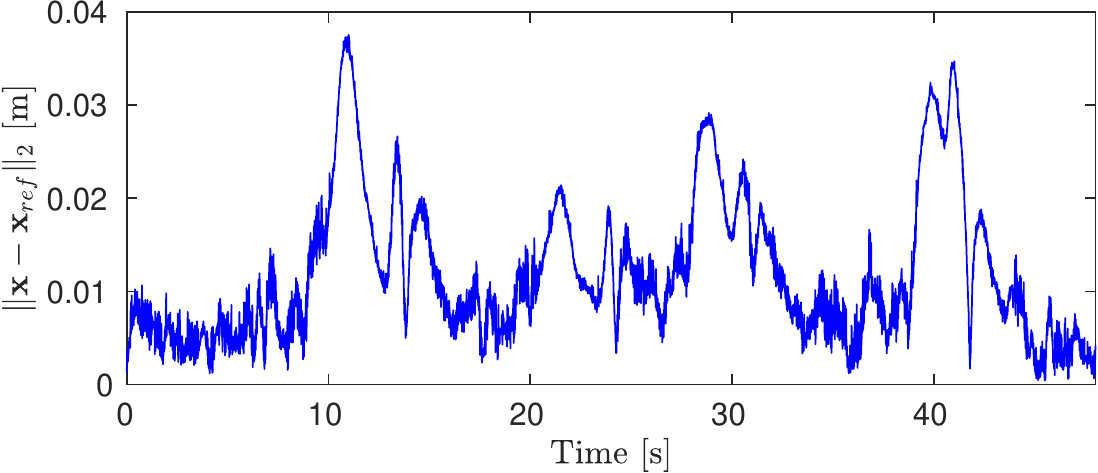}
	\caption{Euclidean norm of position error for hover with disturbance force through tensioned wire.}\label{fig:hover_pos}
\end{figure}

\begin{figure}
	\centering
	\includegraphics[width=\linewidth]{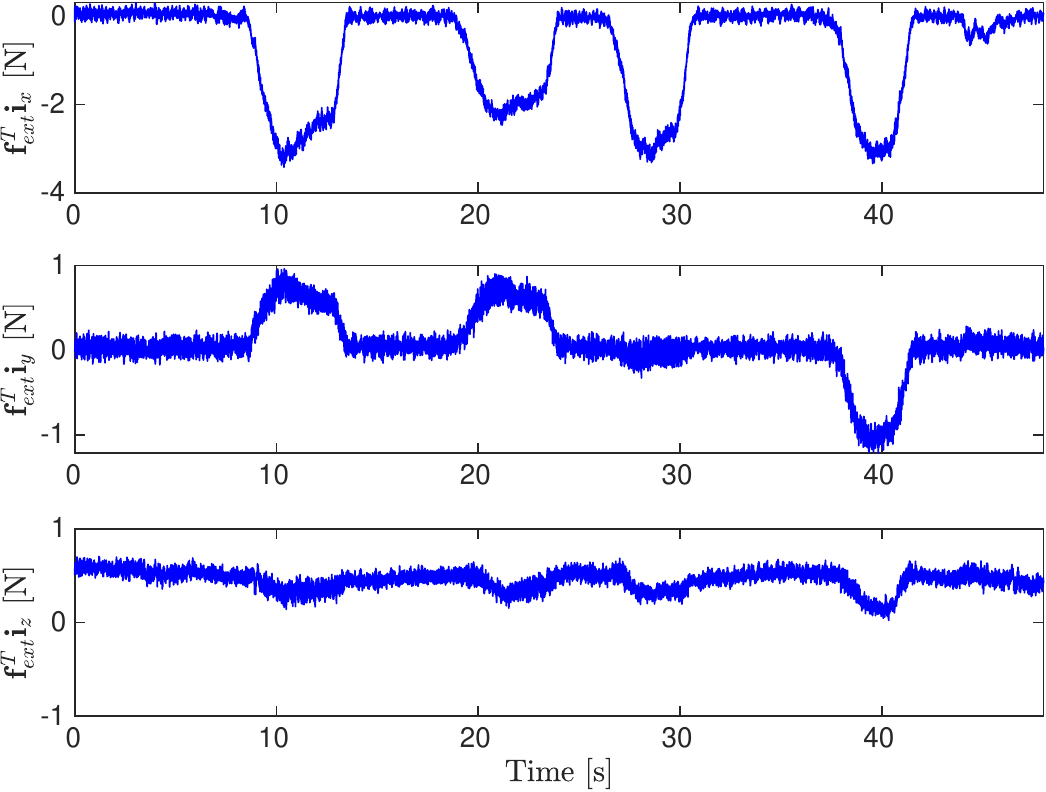}
	\caption{Estimated external disturbance force for hover with disturbance force through tensioned wire.}\label{fig:hover_fext}
\end{figure}

\begin{figure*}
	\centering
	\subfloat[Time is 22 \si{s}.]{%
		\includegraphics[width=0.3\linewidth]{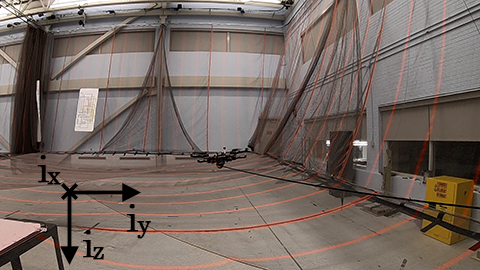}}
	\label{hover5}\hspace{1em}
	\subfloat[Time is 28 \si{s}.]{%
		\includegraphics[width=0.3\linewidth]{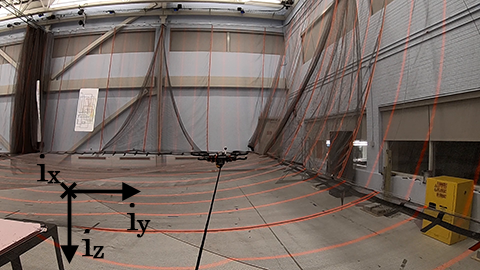}}
	\label{hover20}\hspace{1em}
	\subfloat[Time is 40 \si{s}.]{%
		\includegraphics[width=0.3\linewidth]{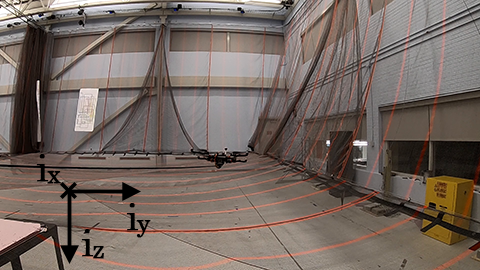}}
	\label{hover25}
	\caption{Quadrotor in hover with disturbance force through tensioned wire.}
	\label{fig:hover_pics} 
\end{figure*}
For a constant $\vect{\sigma}_{ref}$ input, \ie, hover, the controller consistently achieves sub-centimeter position tracking error if no external disturbance is purposely applied. In this section, we present results for hover with an external disturbance force through a tensioned wire. One end of the wire is attached to the bottom plate of the quadrotor. We pull on the other end of the wire to drag the vehicle away from its hover position.

In Fig. \ref{fig:hover_pos}, it can be seen that the quadrotor maintains its position to within at most 4 \si{cm}, while a changing disturbance force is applied through the wire. The largest position error occurs around 10 \si{s} when an external force of approximately 3.7 \si{N} is applied.
Figure \ref{fig:hover_fext} shows the estimated external disturbance force, computed according to \eqref{eq:fext}. The force component in the $\vect{i}_z$-direction has a small steady-state value due to discrepancy between true and estimated thrust. Comparison to Fig. \ref{fig:hover_pics} shows that the direction of the estimated external disturbance force vector corresponds to the direction of the wire. For example, at 22 \si{s}, Fig. \ref{fig:hover_fext} shows that the external force has a negative component in the $\vect{i}_x$-direction and a positive component in the $\vect{i}_y$-direction, and in Fig. \ref{fig:hover_pics}(a) the wire is indeed tensioned in negative $\vect{i}_x$- and positive $\vect{i}_y$-direction.

\section{Conclusions}
In this paper, we proposed a novel control system for the tracking of aggressive, \ie, fast and agile, trajectories for quadrotor vehicles.
Our controller tracks reference position and yaw angle with their derivatives of up to fourth order, specifically, the position, velocity, acceleration, jerk, and snap along with the yaw angle, yaw rate and yaw acceleration using incremental nonlinear dynamic inversion and differential flatness.
The tracking of snap was enabled by closed-loop control of the propeller speeds using optical encoders attached to each motor hub.
The resulting control system achieves 6.6 \si{cm} RMS position tracking error in agile and fast flight, reaching a top speed of 12.9 \si{m/s} and acceleration of 2.1\si{g}, in an 18 \si{m} long, 7 \si{m} wide, and 3 \si{m} tall flight volume.
Our analysis and experiments demonstrated the robustness of the control design against external disturbances, making it particularly suitable for high-speed flight where significant aerodynamic effects occur. 
The proposed controller does not require any modeling or estimation of aerodynamic drag parameters.

\section*{Acknowledgment}
The authors thank Gilhyun Ryou for help with the experiments.

%\addtolength{\textheight}{-4.4cm}   % This command serves to balance the column lengths

\bibliographystyle{IEEEtran}
\bibliography{IEEEabrv,./refs}

\begin{IEEEbiography}[{\includegraphics[width=1in,height=1.25in,clip,keepaspectratio]{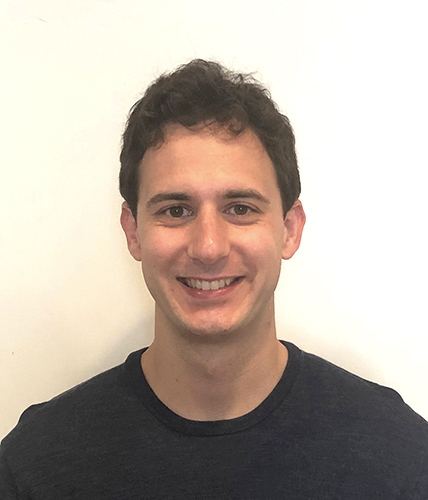}}]{Ezra Tal}
received BSc and MSc degrees from the Faculty of Aerospace Engineering, Delft University of Technology in 2012 and 2015, respectively. He is currently pursuing a PhD degree at the Massachusetts Institute of Technology. In 2012 Ezra was a visiting student at the Technion -- Israel Institute of Technology, and in 2015 he visited NASA Ames Research Center (ARC) as a graduate research intern. He is the recipient of a Huygens Talent Scholarship, and a NASA Group Achievement Award as part of the Adaptive Aeroelastic Wing Shaping Control Team at ARC. His current research interests include differential games and robust control theory, particularly for applications in planning and control of robotics vehicles.
\end{IEEEbiography}

\begin{IEEEbiography}[{\includegraphics[width=1in,height=1.25in,clip,keepaspectratio]{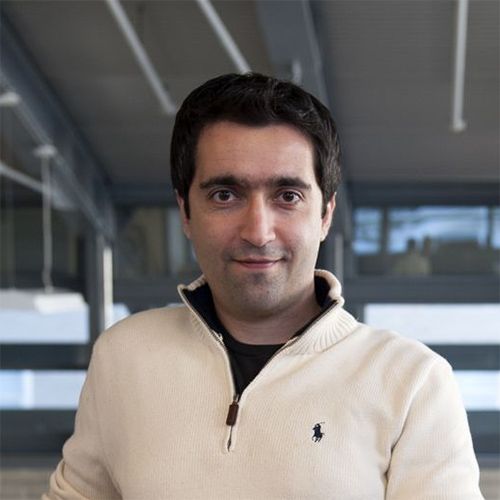}}]{Sertac Karaman} is an Associate Professor of Aeronautics and Astronautics at the Massachusetts Institute of Technology (MIT). He obtained an S.M. degree in mechanical engineering in 2009, and a Ph.D. degree in electrical engineering and computer science in 2012, both from MIT. He studies the applications of probability theory, stochastic processes, stochastic geometry, formal methods, and optimization for the design and analysis of high-performance cyber-physical systems. The application areas of his research include driverless cars, unmanned aerial vehicles, distributed aerial surveillance systems, air traffic control, certification and verification of control systems software, and many others. He is the recipient of an IEEE Robotics and Automation Society Early Career Award in 2017, an Office of Naval Research Young Investigator Award in 2017, Army Research Office Young Investigator Award in 2015, National Science Foundation Faculty Career Development (CAREER) Award in 2014. 
\end{IEEEbiography}

\clearpage

\end{document}